\documentclass{article}

\usepackage[eandd, preprint]{neurips_2026}

\usepackage[utf8]{inputenc}
\usepackage[T1]{fontenc}
\usepackage{hyperref}
\usepackage{url}
\usepackage{booktabs}
\usepackage{array}
\usepackage{enumitem}
\usepackage{amsfonts}
\usepackage{amsmath,amssymb}
\usepackage{nicefrac}
\usepackage{microtype}
\usepackage{xcolor}
\newif\ifshowreviewcolors
\showreviewcolorsfalse  
\let\OriginalTextColor\textcolor
\renewcommand{\textcolor}[2]{%
  \ifshowreviewcolors
    \OriginalTextColor{#1}{#2}%
  \else
    #2%
  \fi
}
\usepackage{graphicx}
\usepackage{verbatim}
\usepackage{soul}
\usepackage{enumitem}
\graphicspath{{figures/}}
\usepackage{subcaption}
\usepackage{multirow}
\usepackage{listings}
\usepackage{pifont}

\title{Rollout Cards: A Reproducibility Standard \\ for Agent Research}

\author{%
  Charlie Masters \\
  Deepflow \\
  \texttt{charlie.masters@deepflow.com} \\
  \And
  Ziyuan Liu \\
  Nanyang Technological University \\
  \And
  Stefano V. Albrecht \\
  Nanyang Technological University \\
}

\begin{document}

\maketitle


\begin{abstract}
Reproducibility problems that have long affected machine learning and reinforcement learning are now surfacing in agent research: papers compare systems by reported scores 
while leaving the rollout records behind those scores difficult to inspect. For agentic tasks, this matters because the same behaviour can receive different reported scores when 
evaluations select different parts of a rollout or apply different reporting rules. In a \textcolor{blue}{structured} audit of 50 popular training and evaluation repositories, we find that none report 
how many runs failed, errored, or were skipped alongside headline scores. We also document 37 cases where reporting rules can change task-success rates, \textcolor{blue}{cost/token accounting, or timing measurements} for \textcolor{blue}{fixed evidence}, 
sometimes dramatically. We treat rollout records, not reported scores, as the unit of reproducibility for agent research. We introduce \emph{rollout cards}: publication bundles that preserve the rollout record and declare the views, 
reporting rules, and drops manifests behind reported scores. We validate rollout cards in two settings. First, four partial public releases in tool safety, multi-agent systems, theorem proving, and search let us compute analyses 
their original reports did not include.
Second, re-grading preserved benchmark outputs across short-answer, code-generation, and tool-use tasks shows that changing only the reporting rule can \textcolor{blue}{change reported scores} by 20.9 absolute percentage points and, in some cases, 
invert rankings of frontier models. We release a reference implementation integrated into \textsc{Ergon}, an open-source reinforcement learning gym, and publicly publish \textcolor{blue}{\textsc{Ergon}-produced} rollout-card exports for benchmarks
spanning tool use, software engineering, web interaction, multi-agent coordination, safety, and search to support future research.

\end{abstract}

\section{Introduction}
\label{sec:intro}
Language-model capabilities have grown faster than the research ecosystem evaluating them. Agent results are compared across a fragmented infrastructure of evaluation harnesses (code that runs tasks, calls models or tools, and scores outputs), RL trainers, agent frameworks, and tracing tools, each with its own conventions for recording rollouts and computing metrics. This is the kind of fragmentation that has historically undermined reproducibility in machine learning and reinforcement learning \citep{henderson2018deeprl, pineau2021improving, agarwal2021precipice}\textcolor{blue}{.}

Early evidence suggests this issue is already surfacing in language-model and agent research. On MMLU~\citep{hendrycks2021mmlu}, identical LLaMA-65B weights score 63.7 under the Berkeley evaluation code and 48.8 under lm-evaluation-harness~\citep{gao2024lmevalharness}: a 14.9-point gap attributable to evaluation-code choices about prompt templates and answer extraction for the same policy~\citep{beeching2023openllm, biderman2024}. Neither pipeline publishes the prompts, model outputs, or grading inputs that would let a third party recompute the score under another reporting rule. The same reporting-rule disagreement appears in public infrastructure records. Production tracing systems disagree on cached-token
accounting~\citep{opentelemetry3163, langfuse12306}, and a public issue in the Verl repository documents a tokenisation-channel divergence reproduced independently by multiple users~\citep{verl2165}.

Across these cases, the missing object is the rollout record behind the reported score, together with the reporting rule that produced it. Current practice publishes the score but discards or under-specifies that record and rule. Later researchers cannot diagnose whether a disputed score gap reflects model behaviour or evaluation-code convention, cannot re-analyse the same runs to study behaviours the original report did not include, and cannot replay another reporting rule against the same outputs. We therefore treat rollout records, rather than reported scores, as the unit of reproducibility for agent research. To substantiate this position, we make three contributions.

\begin{itemize}[leftmargin=*]\itemsep 1pt
\item We diagnose two publication failures that limit reproducibility in current agent evaluation practice (\S\ref{sec:problem}): \emph{recording}, where rollouts are scored once and discarded, and \emph{reporting}, where reporting rules vary without disclosure. \textcolor{blue}{A structured audit of 50 popular training and evaluation repositories documents how failures are surfaced, dropped, or absorbed into reported scores. We also catalogue 37 reporting-rule discrepancies across task success, cost/token accounting, and timing, and show how adjacent communities often rebuild scaffolds or rerun policies to study behaviours the original reports did not preserve.} We find that \textbf{none of the audited repositories report failed, errored, or skipped rollouts alongside headline accuracy or score}.

\item \textbf{We propose \emph{rollout cards}}
  (\S\ref{sec:system}), a minimum-sufficient publication
  specification in the Datasheets \citep{gebru2021datasheets} and
  Model Cards \citep{mitchell2019modelcards} lineage. A rollout card
  preserves the task specification, the agent's actions, and the
  environment dynamics. It also states the reporting rule behind each
  reported score and records what information that rule uses or omits. To
  support adoption and enable independent re-analysis, we release a
  reference implementation in an open-source reinforcement learning gym,
  along with \textcolor{blue}{21 card exports: 17 trace-publication exports plus four analytic/recovered-view non-trace exports}.

  \item We evaluate rollout cards in two settings (\S\ref{sec:validation}). First, reanalysing four partial public releases (GAP, MAESTRO, COPRA miniF2F selected logs, and Tree-of-Thought) computes analyses the original publications did not report. For instance, 20.6\% of agent responses certified safe by GAP made forbidden tool calls anyway, and longer realised search in COPRA's miniF2F logs correlates negatively with proof success rather than positively. Second, holding benchmark artefacts fixed and changing only the reporting rule \textcolor{blue}{changes reported scores} by up to 20.9 percentage points, \textcolor{blue}{swaps GPT-4o and Claude 3.5 Sonnet under different graders on $\tau$-bench, and changes MLE-Bench pass rates from 34.2\% to 13.3\% under medal/pass definitions}. 
  
  Together, the two settings show that publishing only \textcolor{blue}{scores} extracts only part of the value of the experiments papers have already run. 
  Their own rollouts contain findings the \textcolor{blue}{primary works} did not report, and the comparisons they enable cannot distinguish agent behaviour from the rule used to record and report it.

\end{itemize}


\section{Recording vs Reporting in the Current Ecosystem}
\label{sec:problem}

Agent papers usually publish reported scores rather than the rollout
records from which those scores are computed. A rollout is the record
of one episode of agent--environment interaction. For the LLM-agent
setting we focus on, this typically includes the environment
configuration, the observations passed to the agent, the actions it
took (model outputs and tool calls), and the resulting environment
state at each step, along with per-step timing and the episode's
terminal status. We define a \emph{rollout batch} as the collection of
episodes used to compute a \textcolor{blue}{reported score}, and a \emph{view} as the
projection of each rollout onto the fields a particular analysis reads.
A view used for a leaderboard ranking might keep only the final answer
and the gold label, whereas a view used for a safety evaluation might
keep tool call traces and refusal markers.

We define a \emph{reporting rule} as the procedure that turns a viewed
rollout batch into a reported score. A reporting rule typically chains
operations such as a grader scoring outputs, a heuristic deciding how
to handle infrastructure failures, and a rule combining
per-episode scores. Two reported scores can therefore differ because
they use different rollout batches, apply different views to the same
batch, or apply different reporting rules to the same view. By construction, each of
these steps can drop information that a later reader cannot recover
from the \textcolor{blue}{reported score} alone.

Publishing reported scores without the underlying rollout records produces two failure modes. The first, \emph{recording},
concerns the rollout batch itself. When the underlying rollouts are not
preserved, no alternative view or reporting rule can be applied
retrospectively, and questions a later community might ask of the same
episodes go unanswered. The second, \emph{reporting}, concerns the view
and reporting rule together. When these are not declared, readers
cannot tell whether a gap between two \textcolor{blue}{reported scores} reflects
different rollout batches, a different view of the same batch, or a
different reporting rule over the same view. We discuss these two failure
modes in the following subsections.

\subsection{\texorpdfstring{Recording: discarded rollouts foreclose later analyses}{Recording: discarded rollouts foreclose later analyses}}
\label{sec:problem:communities}

The recording problem begins after experiments are complete and results are finalised. A paper can correctly detail the reporting rule used for its own result, yet still leave no way for later readers to ask a different scientific question from the same rollout batch.
Once the underlying rollout records are discarded, later communities cannot inspect the same behaviour through another view or reporting rule.

This loss would be less consequential if subsequent researchers could easily recreate the missing rollout batch.
For LLM agents, however, backfilling a missing analysis faces the dual obstacles of reconstructing the original evaluation setup and paying the cost of producing the rollout batch again.
The setup problem is already substantial. LLM-agent scaffolds couple model, prompt template, tool layer, retry logic, and environment, giving them the implementation sensitivity that has long fragmented empirical RL~\citep{henderson2018deeprl,pineau2021improving,agarwal2021precipice}.
They also add faster implementation churn, where changes in model endpoints, serving precision, tool interfaces, scheduler defaults, or library versions can change the behaviour being measured before a later group can rerun the setup.
The cost problem compounds this reconstruction problem. Even when reconstruction is possible, task horizons frontier agents complete have been doubling every seven months \citep{kwa2025timehorizon}, and total frontier-evaluation cost has risen 3--18$\times$ per year as horizons and reasoning-token budgets outpace per-token efficiency gains \citep{gundlach2026priceofprogress,ord2026agentcosts}.
Backfilling a missing analysis therefore requires both reconstructing a moving policy stack and paying frontier rollout costs again.
For LLM agent work, an unpreserved rollout batch is in practice a lost rollout batch.

This backfilling problem is amplified by the nature of how the community discovers research topics. As models cross capability thresholds, important evaluation questions often become visible only after researchers observe where existing systems fail, coordinate, recover, or behave unexpectedly. Multi-agent coordination, process-level reasoning, and frontier-risk analysis all emerged around capabilities that earlier evaluations were not designed to measure~\citep{tran2025masurvey,cemri2025mast,openai2024o1systemcard,meinke2024scheming,phuong2024dangerouscaps}. As a result, by the time a community identifies a research direction, the rollout batches that could support its analysis have been summarised under another metric and discarded.

The loss starts to matter when different communities want different behavioural slices from similar rollout evidence. SWE-bench made repository-level issue resolution a shared set of environments on which initial results were shared, but later work has duplicated harnesses and regenerated rollout batches to \textcolor{blue}{study agent-computer interfaces, training data, contamination, trajectory quality, cross-language robustness, long-horizon complexity, and failure modes while reimplementing the same underlying scaffolds}~\citep{jimenez2024swebench,yang2024sweagent,swesmith2025,swerebench2025,sweeval2025,multiswebench2025,swebenchpro2025}. The same pattern of fragmented reuse appears in web-agent, tool-use, safety, search, proof, and multi-agent settings, where public artefacts often contain enough behavioural detail for later analysis but papers report only an \textcolor{blue}{outcome label or score}~\citep{webarena2024,browsergym2024,qin2024toolllm,taubench2024,agentharm2025,wang2025ragen,restmcts2024,cemri2025mast}. 

The cost goes beyond duplicated effort, because when rollout records are not preserved, later communities can only ask new questions by regenerating expensive rollout batches. This raises the price of synthesis and comparison, so evidence that could have connected adjacent literatures instead remains locked inside one paper's original reported score. Section~\ref{sec:validation} (RQ1) recovers four such findings from preserved public releases, without re-running anything.

\subsection{Reporting: fragmented rules undermine reproducible comparison}
\label{sec:problem:variance}

The same practice that limits scientific reuse in Section~\ref{sec:problem:communities} also undermines reproducible comparison. 
When papers publish only the scores they choose to report, later readers cannot tell whether a gap between a new result and prior work reflects a difference in agent behaviour or a difference in the view and reporting rule used to compute the score. 
Failure handling, grader design, denominator choice, cost accounting, profiling tool choices, and training-loss calculations can all change a \textcolor{blue}{reported score} for the same underlying behaviour. Recent work on LLM-for-software-engineering reproducibility finds that these rules are rarely fully specified, with recurring gaps in environment specification and versioning~\citep{siddiq2025llm4sereproducibility}.

We test whether reporting-rule fragmentation changes \textcolor{blue}{scores} by auditing 50 popular repositories across agent harnesses, evaluation libraries, benchmarks, and RL stacks. 
At a pinned commit for each repository, we inspect how the implementation handles rule-sensitive events that can change a \textcolor{blue}{reported score}, 
including generation failures, reward or judge failures, unparseable outputs, environment failures, mid-run termination, denominator choices, and training-data filters. 
The audit addresses two questions. First, when such events arise, are they reported alongside the \textcolor{blue}{reported score}, or are they dropped, remapped, or folded into ordinary failures? Second, do reporting-rule choices across popular scaffolds produce different values for rollout returns, cost, timing, and training losses on comparable workloads? Methodological details, including criteria and excluded candidates, are in Appendix~\ref{app:survey}; the reporting-rule discrepancy catalogue is in Appendix~\ref{app:variance-catalogue}.

\textcolor{blue}{The audit finds no audited repository that surfaces failed, errored, or skipped rollouts beside the headline score, and it identifies recurring reporting-rule discrepancies across task-success, cost/token accounting, and timing measurements.} \textbf{None of the 50 repositories reports how often rollouts failed alongside the \textcolor{blue}{accuracy or score}.} In our audit, 11 scored entries silently absorb catastrophic failures into the \textcolor{blue}{reported score}, 31 drop failures with no visible counter, and 9 log failures somewhere but do not surface them beside the \textcolor{blue}{reported score}.\footnote{\textcolor{blue}{Counts are over scored audit entries rather than mutually exclusive repositories; simple-evals is split across two subsets because its SimpleQA and non-SimpleQA paths expose different failure-handling behaviour.}}
From these 50 repositories we additionally identify 37 reporting-rule discrepancies where two codebases produce different task-success, \textcolor{blue}{cost/token, or timing values} for the same model output, API call, \textcolor{blue}{rollout batch, matched training run,} or evaluation case. These include task-success gaps of up to 24.6 percentage points, 2.0$\times$ token-accounting differences, \$14.41 cost differences for the same model family, and 3.1$\times$ runtime gaps on matched hardware. Appendix~\ref{app:variance-catalogue} reports the full set with public documentation, pull requests, or issue threads for each case, and Table~\ref{tab:omnibus} gives six representative examples spanning task success, cost, token accounting, and timing.

\begin{center}
\small
\setlength{\tabcolsep}{5pt}
\captionof{table}{Flagship reporting-rule discrepancies across \textcolor{blue}{the three counted}
families. \textcolor{blue}{Six of the 37 task-success, cost/token, and timing pairs}; the full catalogue with per-pair sources is in
Appendix~\ref{app:variance-catalogue}. \textcolor{blue}{$\Delta$ is the reported gap for the paired evidence in each row; per-pair sources and inclusion criteria are in Appendix~\ref{app:variance-catalogue}.}}
\label{tab:omnibus}
\resizebox{\textwidth}{!}{%
\begin{tabular}{@{}lllll@{}}
\toprule
\textbf{Family} & \textbf{Setup} & \textbf{Convention A} & \textbf{Convention B} & \textbf{$\Delta$} \\
\midrule
Task success  & LLaMA-65B MMLU 5-shot     & Berkeley (63.7)           & lm-eval-harness (48.8)       & \textbf{14.9pp} \\
Task success  & Mistral/Mixtral MMLU      & best prompt template      & worst prompt template        & up to \textbf{24.6pp} \\
Cost/tokens   & Anthropic cached API call & OpenTelemetry input-inclusive & Anthropic separated      & \textbf{2.0$\times$} \\
Cost/tokens   & Sonnet 3.5 on Aider       & Edit benchmark (\$0)      & Refactor+Polyglot (\$14.41)  & \textbf{\$14.41} \\
Latency/timing & SWE-bench Verified       & Docker backend            & Modal backend                & ``systematic'' \\
Latency/timing & GSM8K-GRPO epoch         & optimized TRL             & OpenRLHF                     & \textbf{3.1$\times$} \\
\bottomrule
\end{tabular}
}
\end{center}

We find discrepancies wherever reported scores depend on implementation choices. At the outcome layer, OpenAI Evals maps some invalid judge strings to fixed scores, while Inspect AI surfaces per-sample errors by default~\citep{openai2023evals,aisi2024inspect}. At the measurement layer, OpenTelemetry counts cached Anthropic tokens inclusively while Anthropic's native accounting separates cached reads~\citep{opentelemetry3163}. At the training layer, Verifiers turns reward-function exceptions into zero rewards, while TRL's GRPO path treats missing reward-function outputs as NaNs before group-advantage computation~\citep{brown2025verifiers,trlgrpo2025}. Across these layers, the same output, trajectory, or rollout batch can produce materially different reported scores. This hurts comparison because the \textcolor{blue}{reported score} alone cannot show whether a difference between two systems comes from agent behaviour or from the reporting rule used to record and report that behaviour.

The recording and reporting failures therefore point to the same missing publication object. Papers publish a score, but not the rollout records, rollout batch, view, and reporting rule needed to reuse the evidence or reproduce the comparison. Section~\ref{sec:system} introduces rollout cards as a minimum bundle that preserves the rollout records, declares the rollout batch's view and reporting rule, and records what the reported view leaves out.

\section{Rollout Cards}
\label{sec:system}

Agent evaluations need a publication object that preserves a rollout before a view and reporting rule turn it into a reported score. We call this object a \emph{rollout-card publication bundle}. The \emph{rollout record} stores the episode evidence: task, environment, agent actions, artefacts, timing, status, and failures. The \emph{rollout card} packages that record with a reporting-rule registry, drops manifests, and \textcolor{blue}{release-scope metadata}, which together name the views and rules applied to the record, document what each reported score used or omitted, and \textcolor{blue}{declare any redaction, access, licensing, or redistribution limits}.

We implement this specification in \textsc{Ergon}, an open-source reinforcement learning gym.
\textsc{Ergon} makes rollout-card production a lightweight dataset-adapter which validates a local \textcolor{blue}{artefact}, maps it into the common record, attaches the
available reporting rules and drops manifests, and writes the result as reusable
rollout evidence. We release this implementation to make the specification concrete and provide reusable tooling for researchers
who want to preserve, replay, or re-score agent rollouts after publication. \textsc{Ergon} is not required to use rollout
cards: any framework can emit the same \textcolor{blue}{bundle specification}. 

\subsection{What a rollout card contains}
\label{sec:system:format}

The core component of a rollout card is the rollout record, a self-describing archive readable without access
to the runtime that produced it, so later readers can inspect, re-score,
or re-analyse a completed rollout after publication. In \textsc{Ergon}'s reference schema, the
record contains run metadata; agent or worker nodes; dependency edges;
task and environment state; actions such as messages, model outputs,
tool calls, and tool results; artefacts such as files or reports; and
environment records describing external state changes.
Appendix~\ref{app:system} gives the complete archive layout and row
schemas. A final-answer transcript would preserve only the answer and perhaps
the last model message. A rollout card also records which agent
component acted, what result the environment returned, when failures or
cancellations occurred, and which artefacts downstream graders could
access; failures are status changes in the rollout record rather than
exceptions that bypass logging, addressing the failure-visibility
problem documented in the audit. 

These fields support the community
analyses that current publication records make difficult to reuse:
worker identity and message flow support role-specialisation and
failure-attribution analyses \citep{cemri2025mast, wang2024naht};
turn-level actions and observations support long-horizon analyses of
exploration and recovery \citep{chen2025loop, wang2025ragen}; and
dependency structure, task state, and artefacts support recursive-agent
and hierarchical analyses of delegation, abandoned branches, and final
outputs \citep{bacon2017optioncritic, zhu2024redel}. New analyses can
add domain-specific fields under their own namespaces, so a group
studying multi-agent negotiation can record proposals, acceptances,
refusals, and delegated responsibilities without changing how a
leaderboard view reads the final answer.

\subsection{Views, reporting rules, and drops manifests}
\label{sec:system:drops}

Section~\ref{sec:problem} defined a view as the part of the rollout record
a reported analysis reads, and a reporting rule as the procedure that
turns that view into a reported score. A rollout-card bundle makes
both inspectable. The reporting-rule registry names every view and
reporting rule applied to the card, with its implementation,
configuration, or version where available. A \emph{drops manifest}
accompanies each entry: a typed record of which card streams, fields,
and rows the analysis read; which filters or exclusions it applied;
which structures it collapsed; and which semantic loss classes it
declares for downstream analyses. The complement of this footprint is
what the reported view does not carry forward. \textcolor{blue}{The same mechanism also supports responsible release: when privacy, licensing, scale, or source policy prevents publishing a full trace, the card records the omitted fields or rows, the reason for omission, and whether the remaining artefact is a full trace, redacted trace, gated trace, derived view, or metadata-only record.} In \textsc{Ergon}, each reporting rule is implemented as a versioned scoring
script that accesses cards through a typed reader logging field and row
access. Built-in rules declare their target view and structural
collapses automatically. Custom rules additionally publish their
implementation and any semantic loss labels that field access alone
cannot infer. Field-level claims are therefore replayable from the
card, while semantic labels remain inspectable declarations.

Consider a rollout record that includes how long each environment action took. An episode-return reporting rule may read rewards and terminal status while omitting action-timing and dependency fields. The reported score remains useful for RL, but its drops manifest records losses such as duration and precedence erasure so later readers know that total-plan-time or execution-feasibility analyses require returning to the full card.

\section{Experiments}
\label{sec:validation}

\textcolor{red}{We discussed in Section~\ref{sec:problem} that publishing only reported scores undermines reproducibility by discarding rollout evidence and hiding the impacts of reporting conventions.}
Without the rollout, later research cannot ask new questions of the same behaviour; without the reporting rule,
readers cannot tell whether score gaps reflect model behaviour or reporting-rule construction. \textcolor{red}{In this section, we evaluate experimentally whether rollout cards, as proposed in Section~\ref{sec:system}, address both failures}
\textcolor{red}{Our experiments address two questions:}

\noindent\textbf{RQ1.} When agent papers publish public rollout records,
do those records preserve enough rollout evidence for another research
community to compute a different view of the same episodes?

\smallskip
\noindent\textbf{RQ2.} Holding the rollout evidence fixed, can rollout cards isolate reporting-rule effects, and can those effects be large enough to change the scientific conclusion drawn from the same evidence?

To answer these questions, we build rollout-card exports for public agent-evaluation artefacts spanning tool use, safety, 
software engineering, web interaction, multi-agent coordination, and search. We publicly release \textcolor{blue}{21 card exports: 17 trace-publication exports plus four analytic/recovered-view non-trace exports}
and their dataset adapters so that later researchers can inspect or reanalyse the evidence available for each export class and extend the rollout-card \textcolor{blue}{standard proposal} implemented
in \textsc{Ergon}. Appendix~\ref{app:benchmarks} summarizes the preserved fields, views, reporting rules, and drops for each source; \textcolor{blue}{Appendix~\ref{app:export-storage-profile} reports export size and compression.}

\subsection{Experimental Setup}
\label{sec:validation:questions}
\label{sec:validation:setup}

For RQ1, the test is whether a release built for one \textcolor{blue}{reported score}
contains enough rollout evidence to answer a different community question without
rerunning the policy, environment, or harness. A case qualifies when the release is
publicly available and preserves enough of the rollout record to support adding new reporting rules without repeating rollouts. 
Each recovered view is computed directly from preserved records, using only lightweight offline parsing and aggregation rather than rerunning the policy, environment, or harness.
The results below highlight four releases from the bundle set because they support
analyses their \textcolor{blue}{reported scores} did not report, spanning tool safety, multi-agent
systems, formal theorem proving, and search. We convert each
release into the rollout-card representation from Section~\ref{sec:system},
using preserved rollout fields as the rollout record and recording
missing or source-specific fields as limitations of the recovered view.
Appendix~\ref{app:rq1-reuse} \textcolor{blue}{gives the selection rule, included cases, near misses, source files, extraction rules, and limitations.}

For RQ2, we ask whether rollout cards make \textcolor{red}{reporting-rule}
effects measurable, and whether those effects are large enough to change
reported conclusions.
\textcolor{red}{In each case, we hold the preserved artefact fixed and change only
the reporting rule, measuring the impact of how the same evidence is
extracted, counted, weighted, or judged.}
We sample public archives that can be scored again across a spectrum of
evaluation complexity, from single-completion tasks to
code patches, benchmark submissions, and tool-mediated interactions.
We include a source when its posted view supports at least two existing
public reporting rules over the same rollout evidence, varying answer
extraction, judge heuristics, failure handling, threshold rules, slice
weighting, or trajectory success definitions.
\textcolor{red}{For each case, we therefore report whether the resulting \textcolor{blue}{score
gap} is large enough to change what a reader would conclude from
the same evidence: which model appears better on a benchmark, how large a reported capability
gap is, or whether \textcolor{blue}{a score crosses a benchmark-defined boundary} such
as medal/no-medal.} Both experiments are retrospective: public artefacts test whether evidence released for one reporting rule can support later reuse and re-scoring.

\subsection{Results}
\label{sec:validation:results}

\begin{figure}[t]
\centering
\makebox[\textwidth][c]{%
\includegraphics[width=1.03\textwidth]{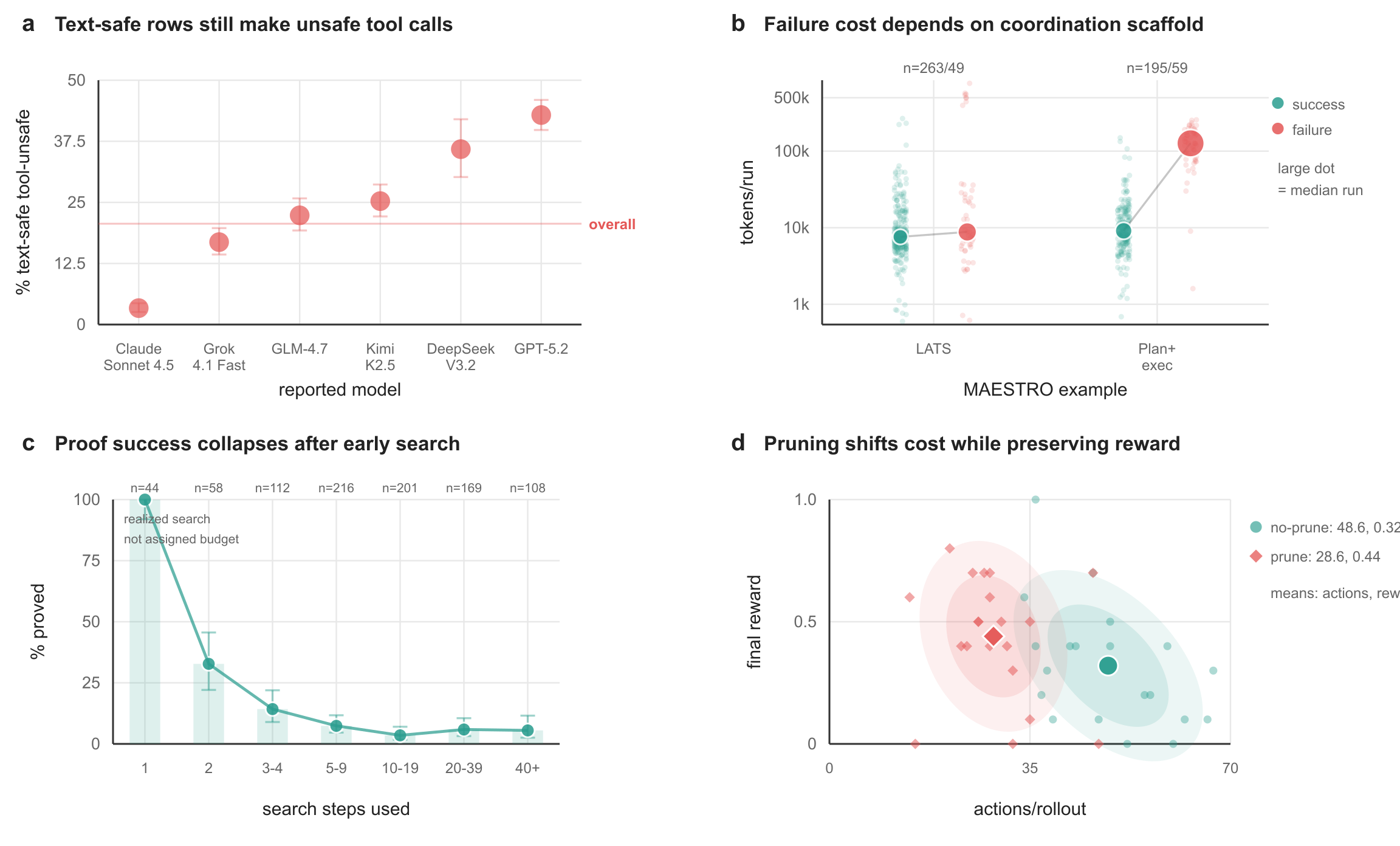}}
\caption{Public rollout releases contain analyses not reported
by their original benchmark scores. \textcolor{red}{Each panel reuses one fixed public release to examine a research question that the original benchmark or paper did not ask:}
(\textbf{a}) GAP text-safety labels versus tool-call safety; (\textbf{b})
MAESTRO run outcome versus coordination overhead; (\textbf{c}) miniF2F
proof outcome in COPRA logs versus realised proof-search cost; and (\textbf{d})
Tree-of-Thought final reward versus action-budget and return profile.}
\label{fig:rq1-recovered}
\end{figure}

\paragraph{RQ1: preserved rollouts support cross-community reanalysis.}
Figure~\ref{fig:rq1-recovered} computes secondary analyses from fixed preserved records. All four panels use
descriptive summaries computed from these records rather than fresh rollouts: model-level rates for GAP \citep{cartagena2026mindgaptextsafety}, run-level medians for MAESTRO \citep{maestro},
binned success rates for miniF2F \citep{zheng2022minif2f} proof attempts from COPRA \citep{thakur2024context} selected logs, and policy-level action/reward means for
Tree-of-Thought \citep{yao2023tree}.

\textbf{Figure 1(a).} The original GAP benchmark \citep{cartagena2026mindgaptextsafety} reports whether a tool-using agent refuses harmful requests in its written response. We ask in a new analysis whether rows judged text-safe, meaning the visible response refuses or avoids harmful content, also avoid forbidden tool calls.
Because the released file contains both labels, our analysis does not require rerunning the policy. Of 4,855 text-safe samples, 1,002 contain unsafe tool calls (20.64\%). Of those 1,002, 997 visibly refuse the harmful request and none expose forbidden content in the text response, so the unsafe behaviour appears only in the tool channel. A response-only safety score can therefore certify visible refusal while missing forbidden actions in the environment.

\textbf{Figure 1(b).} MAESTRO \citep{maestro} uses task success to compare multi-agent software-engineering systems across team structures and orchestration styles.
\textcolor{red}{Aggregating the released runs, we find that failures involve substantially more coordination work than successes.}
\textcolor{blue}{Across directly outcome-labelled runs, failures have median 48 spans and 78{,}523 tokens, while successes have median 10 spans and 11{,}586 tokens.
The median failed run therefore uses roughly 5$\times$ more spans and 7$\times$ more tokens than the median successful run.}
\textcolor{red}{This pattern cuts against the usual test-time-scaling interpretation of multi-agent debate, where extra collaborative work is expected to improve answers \citep{yang2025revisitingmultiagentdebate}. In the released records, extra work \textcolor{blue}{is associated with failure consistent with coordination overhead rather than useful scaling}, a distinction the outcome label alone collapses.}

\textbf{Figure 1(c).} COPRA's \citep{thakur2024context} selected rollouts record attempts to prove olympiad-style formal mathematics problems from miniF2F \citep{zheng2022minif2f}. 
We ask what happens after the prover fails to solve a problem quickly: does additional proof search recover success through further exploration, or does it usually add steps without producing a proof? Across
908 theorem-result records, all 44 one-step attempts succeed, while only
74 of the remaining 864 attempts do. Among attempts longer than one step, success
falls from 27.4\% at 2--3 steps to 7.6\% at 4--7 steps and remains low
thereafter; a logistic model finds that each doubling of realised steps is
associated with lower odds of success (odds ratio 0.62, \textcolor{blue}{$p < 10^{-6}$}).
Because realised search length is observational and the logs do not expose
failed proof states, \textcolor{red}{the result should be read as an association rather than causal evidence that larger budgets cause failure.}
\textcolor{red}{It still exposes a process-level failure mode: repeated proof-search steps can reflect failed recovery rather than useful extra reasoning \citep{gemini2025technicalreport}.}

\textbf{Figure 1(d).} Tree-of-Thought \citep{yao2023tree} records crossword-solving rollouts and reports final word-level reward.
We ask whether this reward hides differences in search efficiency by sampling rollouts which pair depth-first searches on the same 20 puzzles, one with pruning and one without. This lets us stratify reward and action-path measures by pruning condition.
Pruning raises mean final word reward from 0.320 to 0.440 while reducing mean unique atomic actions from 48.65 to 28.65. 
In five pairs, both runs finish with the same final word-level reward despite large differences 
in action coverage, depth, and backtracking. \textcolor{blue}{In this 20-puzzle paired sample, the recovered view is consistent with efficient-search claims by showing that pruning can preserve or improve reward while reducing wasted exploration}, a mechanism final reward alone cannot expose.

Together, the four panels show that the original \textcolor{blue}{reported scores}
answered their papers' native questions, but did not exhaust the evidence in
the released rollouts. \textcolor{blue}{The same records also support adjacent views the original reports did not enumerate.} 
\textcolor{red}{\textcolor{blue}{This illustrates} scientific reuse that Section~\ref{sec:problem:communities} establishes is impeded by only publishing \textcolor{blue}{scores}.}
\textcolor{red}{When releases preserve enough of the rollout record, later communities can ask new questions of the same expensive behaviour rather than rerunning the policy, environment, and harness.}
\textcolor{blue}{The four cases were selected from a broader scout pass (Appendix~\ref{app:rq1-reuse}); most public agent releases preserve only final outputs, which is itself evidence of the recording problem of \S\ref{sec:problem:communities}.}

\paragraph{RQ2: reporting rules can change scientific conclusions.}
After RQ1 shows that preserved rollouts support new views, 
\textcolor{red}{RQ2 asks whether changing only the reporting rule can change the scientific conclusions drawn from fixed evidence.}
We found that this \textcolor{blue}{changes reported scores} by up to 20.9 percentage points and changes model-ordering, \textcolor{blue}{score-gap-size}, and medal-threshold conclusions in our sweep
(Table~\ref{tab:rq2-consequences}), which organizes cases by what the
rollout card holds fixed and which reporting choice changes. Each row in Table~\ref{tab:rq2-consequences} fixes the underlying model,
archived output, or benchmark case and changes only the rule that turns
that artefact into a reported score. These \textcolor{blue}{score gaps} change the conclusions a reader would draw: whether a method improves over a baseline, which model appears better, or how large a claimed capability gap is.

\begin{center}
\scriptsize
\setlength{\tabcolsep}{2pt}
\captionof{table}[Applying alternative reporting rules to fixed public artefacts changes reported scores.]{\textcolor{blue}{Applying alternative reporting rules to fixed public artefacts changes reported scores and, in several cases, the scientific conclusion.
Rows hold fixed the relevant public artefact for each case: generated outputs, submissions, trajectories, or paired public submissions decomposed under denominator conventions.} ``pp'' denotes percentage points.}
\label{tab:rq2-consequences}
\begin{tabular}{@{}>{\raggedright\arraybackslash}p{0.15\linewidth}
                >{\raggedright\arraybackslash}p{0.19\linewidth}
                >{\raggedright\arraybackslash}p{0.20\linewidth}
                >{\raggedright\arraybackslash}p{0.13\linewidth}
                >{\raggedright\arraybackslash}p{0.20\linewidth}@{}}
\toprule
\textbf{Benchmark} & \textbf{Held fixed} & \textbf{Reporting choice}
& \textbf{\textcolor{blue}{Score gap}} & \textbf{Consequence} \\
\midrule
HumanEval/GPQA & Generated outputs & Driver/extraction variants
& 0.6pp/1.0pp & Minor changes: independent rules mostly agree \\
BrowseComp & Gold/predicted answers & Rule-based vs LLM judge
& 4.17pp & Judge convention changes 274 answer labels \\
SWE-bench Verified & GPT-4o submissions & Missing patches counted as failures vs excluded& 2.3pp of 15.6pp gap & Affects reported capability gap \\
\(\tau\)-bench & Tool-call trajectories & DB-state vs sequence/set graders
& 16.9pp & 12 changes in ordering of frontier models \\
MLE-Bench & Kaggle submissions & Medal/pass definition
& 20.9pp & Gold-only reporting cuts pass rate from 34.2\% to 13.3\% \\
\bottomrule
\end{tabular}
\end{center}

On SWE-bench Verified \citep{jimenez2024swebench}, both public
submissions use GPT-4o-2024-05-13, but no-submission rates differ, with
50/500 for SWE-agent \citep{yang2024sweagent} and 4/500 for Agentless
\citep{xia2024agentless}. Counting no-submissions as unresolved gives a
15.6pp gap; scoring only submitted patches gives 13.3pp, so the
reporting rule accounts for 2.3pp of the \textcolor{blue}{reported score gap}. Other rows show
larger rule effects. \(\tau\)-bench grader choice \textcolor{blue}{changes reported scores} by
16.9pp and reverses the GPT-4o versus Claude 3.5 Sonnet ordering, while for MLE-Bench the same archived
Kaggle submissions score 34.2\% under an above-median rule but 13.3\%
under a gold-only medal rule.
\paragraph{Reporting-rule effects grow with evaluation complexity.}
Reporting sensitivity is uneven across evaluation types. HumanEval and
GPQA move only 0.6pp and 1.0pp under independent driver or extraction
variants, giving near-null calibration cases. In agentic settings, the
evaluated object is often a multi-step artefact such as
a software patch, a Kaggle submission, or a tool-call trajectory. The
reporting layer must then choose what counts as an attempted run, which
artefact slice to score, how to weight partial work, and which trajectory
state defines success.
BrowseComp isolates one stochastic layer, with four graders spanning 4.17pp and
disagreeing on 274 of 5,064 preserved answers, including 26 exact-match
judge failures. As evaluations move toward longer-horizon tasks and richer environments, two harnesses can report the same nominal score while making more local choices about what to count, drop, combine, or treat as success. The risk of reporting-rule \textcolor{blue}{score gaps} therefore grows.

\textcolor{blue}{Together, the two experiments show how rollout cards help address} the recording and reporting failures that Section~\ref{sec:problem} identified as an emerging reproducibility problem for agent evaluation. \textcolor{blue}{RQ1 demonstrates that preserved rollouts can mitigate the recording problem by enabling new views over fixed public records.} RQ2 shows that declared reporting rules \textcolor{blue}{mitigate} the reporting problem by making convention-driven score changes inspectable rather than implicit.

\section{Related Work}
\label{sec:related}

Rollout cards extend documentation standards for machine learning \textcolor{blue}{artefacts}. Datasheets for Datasets \citep{gebru2021datasheets} document the motivation, composition, collection, uses, distribution, and maintenance of datasets. Model Cards
\citep{mitchell2019modelcards} document trained models, intended use, evaluation procedures, and performance characteristics. By contrast, rollout cards document completed agent runs and the views and reporting rules used to turn those runs into
\textcolor{blue}{reported scores}.

Prior work standardises parts of the reporting layer. HELM broadens
language-model evaluation across scenarios and measures
\citep{liang2023helm}, while OLMES and Unitxt make evaluation
transforms more explicit and reusable.
\textcolor{blue}{Biderman et al. show prompt-template effects, and Dr.\ GRPO isolates loss-normalisation effects} \citep{biderman2024,liu2025drgrpo}. These
efforts expose reporting rules; rollout cards target the
prerequisite record layer by preserving the rollout record and the
drops manifests needed to replay, compare, or replace those reporting rules
after publication.

Recent surveys~\citep{yehudai2025agenteval, cemri2025mast,
yang2025agentprotocols} flag fragmentation across agent trajectory
formats, and systems such as AgentOhana, VerlTool, and Agent
Lightning~\citep{zhang2024agentohana, jiang2025verltool,
luo2025agentlightning} unify trajectories or agent logic for training.
Those proposals make trajectories easier to consume during training;
rollout cards make completed runs easier to inspect after publication.

\section{Conclusion}
\label{sec:conclusion}

Reproducibility problems that have long affected machine learning and reinforcement learning are now appearing in agent evaluation, where reported scores depend on grading scripts, failure-handling rules, timing windows, and loss calculations. Our audit makes this prevalence visible: across 50 popular repositories and 37 reporting-rule discrepancies, failures and reporting choices are routinely hidden behind headline scores. Rollout cards \textcolor{blue}{help address} this early by preserving the rollout record behind a reported score and pairing it with explicit views, reporting rules, and drops manifests. In our experiments, preserved traces reveal safety, coordination, proof-search, and search-efficiency findings that original reports did not include, while fixed benchmark artefacts can be re-graded to \textcolor{blue}{change reported scores} and flip model orderings.

\textcolor{blue}{\textbf{Limitations.}} Rollout cards do not prevent selective metric choice, privacy constraints, redaction, or selective reporting. Their role is narrower: to make the record, rule, and omissions inspectable so disagreements can be traced to preserved evidence, reporting choices, or agent behaviour. At scale, shared rollout-card repositories would support reuse, reporting-rule comparison, and meta-analyses without new frontier rollout budgets.
\bibliographystyle{plainnat}
\bibliography{references}

\appendix


\section{RQ1 Public-Rollout Reuse Inventory}
\label{app:rq1-reuse}

This appendix documents how we selected the four RQ1 reuse cases and
what each recovered rollout-card representation preserves. The goal is
not to claim an exhaustive survey of all public agent artefacts. It is to
show that the main-text cases were chosen under a stated public-record
criterion rather than after rerunning policies or selecting only
successful examples.

\paragraph{Inclusion rule.}
A candidate had to satisfy four conditions: (i) the release was public
or open source; (ii) it preserved \textcolor{blue}{labels, logs, traces, or process fields} beyond the
\textcolor{blue}{reported score}; (iii) the original paper or benchmark reported a
different primary view; and (iv) the reanalysis required no new policy
rollouts. Offline computation or judging was allowed only when applied
to preserved records and declared as part of the recovered view. 
\textcolor{blue}{The artefact does not redistribute the full upstream source datasets for these cases. Instead, the submitted artefact records derived card records, adapters, manifests, view definitions, reporting-rule metadata, and drops manifests at the level declared in the released metadata. Readers who need the original task inputs or upstream dataset files should obtain them from the original releases.}

\begin{table}[h]
\centering
\small
\setlength{\tabcolsep}{4pt}
\caption{RQ1 public rollout reuse cases. Each included case preserves a
\textcolor{blue}{public record} rich enough to support a second view over the same
behaviour.}
\label{tab:rq1-inventory}
\begin{tabular}{@{}p{0.13\linewidth}p{0.20\linewidth}p{0.21\linewidth}p{0.20\linewidth}p{0.17\linewidth}@{}}
\toprule
\textbf{Case} & \textbf{Source artefact} & \textbf{Original view} &
\textbf{Recovered view} & \textbf{Main limitation} \\
\midrule
GAP & Public results parquet, 17{,}420 rows and 42 columns
\citep{cartagena2026mindgaptextsafety} & Text-safety and refusal labels
& Tool-call safety, forbidden-call fields, and refusal/tool divergence
& Benchmark evidence of reporting-rule non-equivalence, not deployment
prevalence \\
MAESTRO & Public trace parquet, 116{,}918 rows and 1{,}056 runs
\citep{maestro} & Run outcome or judgement & Span/token overhead and
coordination-process structure & Overhead is observational and not a
causal failure mechanism \\
COPRA miniF2F selected logs & Two selected-result logs, 908 theorem-result records
\citep{thakur2024context,zheng2022minif2f} & Proved/failed theorem outcome &
Realised proof-search steps, elapsed time, and successful proof style &
Failed tactic branches and failed proof states are not exposed \\
Tree-of-Thought & Paired crossword DFS prune/no-prune logs, 20 puzzle
pairs \citep{yao2023tree} & Final crossword reward & Action coverage,
depth, backtracking, and prune/no-prune return profile & Small paired
sample; pruning mechanism is reconstructed from traces \\
\bottomrule
\end{tabular}
\end{table}

\paragraph{Extraction rules.}
For GAP, the text-safety view is the released \texttt{t\_safe} field
and the recovered process view uses \texttt{tc\_safe},
\texttt{gap}, \texttt{forbidden\_calls}, refusal strength, and
tool-call counts. The headline main-text statistic is computed over rows
with \texttt{t\_safe=True}: 1{,}002 of 4{,}855 are tool-call unsafe.
For MAESTRO, rows are grouped by \texttt{run\_id}; direct labels use
\texttt{run.outcome} when present, and the recovered overhead view sums
span counts and token fields by run. We report direct outcome labels in
the main text and use task-specific derived labels only as a robustness
check. For the COPRA miniF2F logs, each theorem-result record contributes its
\texttt{SearchResult}, \texttt{StepsUsed}, elapsed search time, and any
successful proof text; binned success rates use realised steps, not
assigned compute budget. For Tree-of-Thought, each DFS record stores
the current action path and reward snapshot. We reconstruct unique
atomic actions, maximum depth, and backtracking from action-list prefixes
and compare paired prune/no-prune traces for the same crossword prompt.

\paragraph{Candidate inventory and near misses.}
The broader scout pass considered public trace artefacts across tool
safety, multi-agent systems, theorem proving, search, planning/UI, web,
coding agents, and function calling. Several candidates were usable but
not selected for the four-panel main-text figure because they overlapped
the final domains or produced weaker source pairings: AgentHarm,
\(\tau\)-bench historical trajectories, SWE-smith trajectories,
agent-reward-bench, ATBench,
StableToolBench, BFCL, MiniWoB++, AgentNet, and Tree-of-Thought Game24.
We excluded other candidates because we could not verify a reusable
public corpus of raw traces plus outcomes, the release was gated or too
large for the submission-time audit, or the available artefact had only
final outputs rather than process logs. These near misses include
standalone ChatDev, MetaGPT, AgentVerse, AutoGenBench, TraceElephant,
LeanDojo/HOList, DeepSeek-Prover/miniF2F final proofs, ReST-MCTS,
rStar-Math, RAGEN, OSWorld, AppWorld, WorkArena, TravelPlanner, ToolQA,
AgentDojo, ToolEmu, OS-Harm, ToolShield, SafeArena, BrowserART, and
\texttt{cx-cmu/agent\_trajectories}. This inventory is why the main text
describes the four panels as a diverse demonstration set rather than a
prevalence estimate over all public agent releases.

\paragraph{Missing fields and recovered-card limits.}
None of the four source releases was originally published as a
rollout-card bundle. The recovered cards therefore preserve only fields
present in the public artefact. GAP lacks the full runtime environment
behind each tool interaction; MAESTRO exposes span and metadata fields
but not a single causal label for coordination failure; the COPRA miniF2F logs expose
theorem-level search records and successful proofs but not failed
tactic-branch states; and the Tree-of-Thought crossword logs expose
action paths and reward snapshots but not every evaluator decision that
led to pruning. These omissions are recorded as recovered-view
limitations rather than filled in by rerunning the policies.


\section{50-repo Failure-Handling Survey}
\label{app:survey}

This appendix reports the code-level evidence underlying the
failure-handling audit in Sec.~\ref{sec:problem}. We shallow-cloned each
repository at a pinned commit SHA in April~2026 and audited it against a
seven-pattern rubric. Every
claim below is supported by a commit-pinned public source reference;
the full references and evidence notes are not reproduced inline for
reasons of space, but are released in
\texttt{ergon\_survey\_supplement/survey\_audit/01\_repository\_failure\_audit/}.
Each record links the paper label to a pinned ref and colocated
evidence note.
Pinned SHAs are given in \S\ref{app:survey:shas}.

\subsection{Repository selection}
\label{app:survey:selection}

\paragraph{Sampling frame and operational scope filter.}
The survey uses \emph{category-stratified purposive sampling} over
repositories that produce benchmark numbers for LLM-agent papers. A
repository is in-scope if and
only if, at its pinned SHA, it contains code that (a)~is
actively maintained (last commit within the past 12 months),
(b)~produces a benchmark number, training-pipeline metric, or
leaderboard rank that a published paper would cite, and
(c)~implements the result aggregation itself rather than
delegating all aggregation to a separately-versioned downstream
harness. Star-count was not used as a threshold; verified
in-scope repositories range from $\sim\!150$ stars (SciCode
164, MLAgentBench 320) to $\sim\!40$k+ (FastChat, aider),
demonstrating that the in-scope/canonical-harness filter
dominates any star-based floor. Operational scope is enforced
by documented scope decisions rather than concealed filtering.
For example, OpenHands was spot-checked
(SHA \texttt{3b17f27}) and documented OUT-OF-SCOPE because its
evaluation harness lives in a separately-versioned repository
(\texttt{github.com/OpenHands/benchmarks}) and the main repo
contains no result-aggregation, scoring, or denominator-handling
code. This is the operational scope filter working as intended:
OpenHands is the foundation layer (analogous to how PyTorch is
the foundation for training frameworks) and is correctly
excluded by criterion~(c).

\paragraph{Categories.}
\emph{(i)} SWE-bench-family harnesses and agent scaffolds
(SWE-bench, SWE-agent, mini-swe-agent, live-swe-agent,
SWE-bench\_Pro-os, SWE-smith, Agentless, aider);
\emph{(ii)}~RL training frameworks (TRL, verl, OpenRLHF, rllm,
slime, ART, Agent-R1, RAGEN, MARTI, MATPO, RL-Factory,
Trinity-RFT, ms-swift, verifiers, open-r1); \emph{(iii)}~general
eval harnesses and LLM-as-judge leaderboards (openai/evals,
simple-evals, SciCode, lm-evaluation-harness, lighteval,
inspect\_ai, FastChat/MT-Bench + Chatbot Arena, HELM, ragas,
deepeval, MTEB, promptfoo, BIG-bench); \emph{(iv)}~web/GUI,
mobile, and ML-engineering / scientific-agent benchmarks
(WebArena, VisualWebArena, OSWorld, android\_world, Mind2Web,
MLE-bench, MLAgentBench, ScienceAgentBench); and \emph{(v)}~%
function-calling and general multi-agent / scaffold repositories
(BFCL/gorilla, ToolBench, AgentBench, AgentBoard, tau-bench,
Self-rewarding-reasoning-LLM). The final sample is 50 in-scope
repositories, with excluded candidates documented as scope-filter
decisions rather than concealed.

\paragraph{Sensitivity to the sampling frame.}
The 50-repository result is therefore a claim about repositories that
implement result aggregation for a benchmark, training metric, or
leaderboard number. It is not a prevalence estimate over every public
trajectory dataset or agent artefact release. To check this boundary, we
maintained a broader discovery log of public agent-artefact releases
while building the rollout-card corpus. After ordinary URL triage, that
log contained 91 candidate releases; a deterministic metadata or sample
fetch yielded public metadata or sample files for 89 releases, with two
gated releases retained as gated. Most releases outside the
50-repository audit were excluded not because they contradicted the
finding, but because they changed the estimand: they were trajectory
datasets, model/checkpoint pages, demonstration corpora, gated releases,
or framework examples rather than repositories whose own code computes
a paper-citable score. Including them would require a separate
``public-release documentation'' audit, not relabelling them as clean
failure-handling implementations. Table~\ref{tab:excluded-candidates}
gives representative excluded candidates and the exclusion rule they
triggered.

\begin{table}[h]
\centering
\small
\setlength{\tabcolsep}{4pt}
\caption{Representative candidates considered but excluded from the
50-repository failure-handling audit. The table records scope decisions,
not negative judgements about the usefulness of the releases. Several of
these artefacts are used elsewhere in the paper or public card release,
but they are not counted in the repository-level failure-accounting
denominator because they do not themselves implement the aggregation
code being audited.}
\label{tab:excluded-candidates}
\begin{tabular}{@{}p{0.25\linewidth}p{0.23\linewidth}p{0.44\linewidth}@{}}
\toprule
\textbf{Candidate} & \textbf{Release type} & \textbf{Reason excluded from the 50-repo audit} \\
\midrule
OpenHands main repository & Agent platform & Spot check found the evaluation harness in a separately versioned benchmarks repository; the main repo did not itself implement the result aggregation code. \\
SWE-smith trajectories; SWE-agent / SWE-rebench trajectory mirrors & Trajectory datasets & Public trajectory releases, not repositories whose own code turns failures into a reported benchmark number. \\
AgentHarm, ATBench, WebLINX & Benchmark or trace datasets & Useful public records for rollout-card exports, but the dataset cards are not the aggregation implementations audited for failure accounting. \\
Agent-FLAN, xLAM, Toucan, ToolAlpaca & Tool-use instruction or training data & Dataset/model releases without a repo-local denominator or failure-handling path for a headline score. \\
TRAIL and xLAM-function-calling & Gated Hugging Face releases & Dataset access was gated during the discovery pass, so they could not be audited against pinned public source lines. \\
Open X-Embodiment, LeRobot collection, DROID, RH20T & Embodied-agent demonstration corpora & Very large robot-demonstration releases outside the LLM-agent aggregation-code scope. \\
Generative Agents, ChatDev, AutoGen examples & Framework or simulation artefacts & Public scaffolds or example logs, but not in-scope aggregation repositories under criterion~(c). \\
\bottomrule
\end{tabular}
\end{table}

\subsection{Audit methodology}
\label{app:survey:method}

We shallow-cloned each public repository during the audit window and
pinned it with \texttt{git rev-parse HEAD}. The 50 SHAs are listed in
\S\ref{app:survey:shas} and used as the source baseline for every
permalink. Every citation in this appendix
resolves to
\texttt{github.com/\{owner\}/\{repo\}/blob/\{SHA\}/\{path\}\#L\{start\}-L\{end\}}
against that SHA. We accepted claims only when the cited
source lines directly exhibited the denominator, failure-handling,
or aggregation behaviour being described. \textcolor{blue}{The audit is
source-verifiable: each label resolves to pinned code that readers can
inspect directly.} The uploaded supplement
records the final repository labels, pinned refs, and evidence IDs;
representative per-repo evidence notes are colocated with those records
to illustrate the template.

\subsection{Severity rubric: seven patterns of silent drop}
\label{app:survey:rubric}

We scored every repository 0--3 against a rubric defined by
seven recurring code patterns:
\emph{(P1)} \texttt{None}-to-\texttt{0.0} reward coercion
(reward-function exception silently yields a numeric zero that
enters aggregation alongside genuine zero-reward rollouts);
\emph{(P2)} variance-based rollout filters that discard entire
rollout groups with zero std (collapse with the remaining group
in published aggregates); \emph{(P3)}
save-on-clean-exit patterns bypassed by
\texttt{KeyboardInterrupt} or early-exit cost-killers, removing
instances from the submission file rather than the denominator;
\emph{(P4)} survivor denominators --- reporting
\texttt{sum(scores) / len(scores)} over instances that reached
the aggregation step, with the pre-aggregation drop count not
preserved; \emph{(P5)} multi-turn N-destroys-1-to-N-1 patterns,
where a mid-rollout tool-call exception discards the entire
trajectory including its already-completed turns; \emph{(P6)}
LLM-judge regex-miss-to-fixed-score, where a judge returning
an unparseable response is coerced to a specific numeric score
indistinguishable from a legitimate one (e.g.\ \texttt{0.0},
\texttt{min(choice\_scores)}, \texttt{NOT\_ATTEMPTED}); and
\emph{(P7)} outcome-as-ground-truth SFT filtering, where a
downstream supervised dataset is built from only those
trajectories that scored correctly on the biased harness,
propagating the bias across a model-generation boundary. A
repository scores 3 if a single pattern is catastrophic and
observable at the headline output (the failure changes the
number a paper would cite); 2 if one or more patterns are
present but the reader cannot tell from the published CSV; 1 if
failures are logged but excluded from the denominator with no
per-category counter; 0 if all failures are surfaced alongside
metrics. No repository in the survey defaults to score 0.

\subsection{Score distribution}
\label{app:survey:distribution}

\begin{table}[h]
\centering
\small
\caption{50-repo severity scores. No repository defaults to
surfacing failures alongside metrics. OpenHands is documented
OUT-OF-SCOPE (no in-repo aggregation harness; benchmarks live
in a separately-versioned repository) and not counted in the
50. Simple-evals appears in both score-3 (SimpleQA subset) and
score-1 (non-SimpleQA subset); the $11{+}31{+}9{=}51$ listings
correspond to 50 unique repositories.}
\label{tab:survey:distribution}
\begin{tabular}{@{}clp{0.55\linewidth}@{}}
\toprule
\textbf{Score} & \textbf{Count} & \textbf{Repositories} \\
\midrule
3 & 11 & SWE-bench, openai/evals, simple-evals (SimpleQA),
VisualWebArena, ToolBench, Self-rewarding-reasoning-LLM,
PrimeIntellect verifiers, Trinity-RFT, ms-swift (GRPO), BFCL
(gorilla), FastChat (MT-Bench + Chatbot Arena) \\
\addlinespace
2 & 31 & Agent-R1, MARTI, SWE-bench\_Pro-os, RAGEN, OpenRLHF,
TRL, ART, slime, tau-bench, rllm, live-swe-agent, SWE-agent,
mini-swe-agent, MATPO, verl, AgentBoard, AgentBench,
RL-Factory, SWE-smith, lm-evaluation-harness, WebArena,
MLAgentBench, MLE-bench, lighteval, Agentless, aider, Mind2Web,
open-r1, ragas, MTEB, promptfoo \\
\addlinespace
1 & 9 & SciCode, inspect\_ai, ScienceAgentBench, android\_world,
OSWorld (harness), simple-evals (non-SimpleQA subset), HELM,
BIG-bench, deepeval \\
\addlinespace
0 & 0 & --- \\
\bottomrule
\end{tabular}
\end{table}

Inspect AI (score 1) is the near-positive control referenced in
Sec.~\ref{sec:problem}: its default
\texttt{fail\_on\_error=True} surfaces per-sample errors, but
the commonly-used \texttt{fail\_on\_error=False} opt-in
degrades silently with no per-category counter
(\texttt{inspect\_ai/\_eval/task/error.py:L26} and
\texttt{inspect\_ai/scorer/\_metrics/accuracy.py:L33} at SHA
\texttt{36231d6}).

\subsection{Per-repo entries}
\label{app:survey:entries}

We summarise each repository's score with a one-line evidence
pointer; the eleven score-3 repositories receive a short
2--3-sentence writeup naming the specific code site(s) and
user-visible consequence. All paths are relative to the repo
root at the pinned SHA (\S\ref{app:survey:shas}). Full source
references and supporting evidence notes are in
\texttt{ergon\_survey\_supplement/survey\_audit/01\_repository\_failure\_audit/}.

\paragraph{Score-3 repositories (11).}

\textbf{SWE-bench.} The grading loop aggregates over
\texttt{instances\_to\_run}, filtered before aggregation
(\texttt{swebench/harness/run\_evaluation.py:L465-L470}), while
empty patches are documented as filtered
pre-submission (\texttt{docs/faq.md:L39}). A container-level
infrastructure failure stays in the denominator as a fail, but
an empty-patch instance is removed from the denominator
altogether --- and no output field separates the two cases in
the published submission.

\textbf{openai/evals.}
\texttt{backoff.on\_predicate} at
\texttt{evals/utils/api\_utils.py:L10-L22} retries transient
predicates indefinitely with no \texttt{max\_tries}, so long
runs silently absorb retry-exhaust events rather than surfacing
them. The MMMU eval treats errors as wrong answers
(\texttt{elsuite/mmmu/eval.py:L158}) and the modelgraded
classifier coerces \texttt{INVALID\_STR} to
\texttt{min(choice\_scores)}
(\texttt{elsuite/modelgraded/classify\_utils.py:L99-L101}):
three distinct P6 patterns in one library.

\textbf{simple-evals (SimpleQA subset).} The regex
fallback at \texttt{simpleqa\_eval.py:L126} defaults unparseable
responses to \texttt{NOT\_ATTEMPTED}, which the \textcolor{blue}{reported score}
(\texttt{accuracy\_given\_attempted = correct / (correct +
incorrect)}) then structurally excludes
(\texttt{simpleqa\_eval.py:L169-L176}). A model that hedges on
every ambiguous question looks arbitrarily accurate under this
metric --- NOT\_ATTEMPTED tasks do not appear in the
denominator.

\textbf{VisualWebArena.} An \texttt{assert "correct"
in response} at
\texttt{browser\_env/helper\_functions.py:L606} raises on
unparseable judge output, causing the outer try/except at
\texttt{run.py:L453-L461} to drop the entire task. Image-fetch
errors at \texttt{image\_utils.py:L44} propagate through the
same path. Dropped tasks are absent from the output, not
counted as failed.

\textbf{ToolBench (9/9 \textsc{verified}).} The DFS
search code at \texttt{DFS.py:L84-L91} randomly promotes a
fraction of \texttt{give\_up\_node} trajectories to
\texttt{valid\_data=True}, and SFT preprocessing at
\texttt{preprocess/preprocess\_toolllama\_data.py:L44-L45}
filters on \texttt{valid\_data} alone --- so ToolLLaMA trains
on the promoted failures. This is the paper's paradigm example
of pattern~P7 (outcome-as-ground-truth SFT).

\textbf{Self-rewarding-reasoning-LLM (4/4
\textsc{verified}).}
\texttt{infer\_math/}\allowbreak\texttt{reward\_labeling.py:L1612-L1622} grants
correctness against the ground-truth answer on the
\emph{initial} chain-of-thought; self-correction attempts must
match that \texttt{first\_reward} to be counted
(\texttt{process\_prompt\_turn3.py:L24-L62}), with a hard cap
$N=3$. There is no \texttt{try/except} anywhere in the
generation module (\texttt{gen\_hf.py}): any infra failure
silently removes a rollout.

\textbf{PrimeIntellect verifiers (5/5
\textsc{verified}).} Reward-function exceptions at
\texttt{rubric.py:L144-L158} yield \texttt{0.0}, and group
reward-function exceptions at \texttt{rubric.py:L208,~L217}
yield \texttt{[0.0]*N}. The errored zeros are then included in
the group-mean advantage at \texttt{rubric.py:L325-L333}
(\texttt{avg\_reward = sum(aggregated\_rewards) / num\_states};
\texttt{t["advantage"] = state["advantage"]}), so an
infrastructure error in one rollout biases the policy gradient
for the other rollouts in its group.

\textbf{Trinity-RFT.}
\texttt{asyncio.TimeoutError} at
\texttt{trinity/explorer/scheduler.py:L214-L216} is packaged
into \texttt{Status(ok=False,~metrics=list(),~message=\ldots)},
but the consumer path at \texttt{scheduler.py:L532-L602}
(\texttt{get\_results}) never inspects \texttt{status.ok} ---
errored payloads flow into training with \texttt{ok=False}
undetected. The over-rollout cancellation grace period
(\texttt{scheduler.py:L559-L572}, 30\,s default from
\texttt{config.py:L147}) means timeouts can also truncate
sub-tasks mid-run (\texttt{scheduler.py:L490-L506}) without
surfacing the partial-completion to the rollout log.

\textbf{ms-swift (GRPO) (8/8 \textsc{verified}).}
None-rewards are lifted to NaN at
\texttt{swift/rlhf\_trainers/grpo\_trainer.py:L359, L367,
L379}, then aggregated with \texttt{.nansum(dim=1)}
(\texttt{grpo\_trainer.py:L473}); DAPO's
\texttt{max\_resample\_times} revert (\texttt{grpo\_trainer.py:%
L684-L733}) and \texttt{overlong\_filter} mask-zeroing
(\texttt{grpo\_trainer.py:L1123-L1129}) both silently remove
rollouts from the gradient. The LLM-judge path
(\texttt{rewards/rm\_plugin.py:L216-L226}) coerces regex-parse
failures to \texttt{0.0} (pattern P6).

\textbf{BFCL / gorilla.} The weighted overall
accuracy formula in
\path{berkeley-function-call-leaderboard/bfcl_eval/eval_checker/eval_runner_helper.py:L509-L519} applies
$[10,10,10,30,40]$ weights over category runners that each
silently decrement \texttt{correct\_count} on parse failure
while leaving the denominator \texttt{len(model\_result)}
unchanged (per-category dispatch at \texttt{eval\_runner.py:%
L668}). \texttt{eval(func\_call)} exceptions become literal
\texttt{"Error during execution: \ldots"} strings
(\texttt{multi\_turn\_eval/multi\_turn\_utils.py:L97-L98}),
scored as a failed call without category attribution.

\textbf{FastChat (MT-Bench + Chatbot Arena).} FastChat is the
reference implementation for two of the most-cited public LLM
evaluation numbers: MT-Bench (LLM-as-judge single-turn and
multi-turn scoring) and Chatbot Arena (pairwise Elo ranking). A
GPT-4 or Claude judge whose response fails to match the
\texttt{[[score]]} or \texttt{[score]} regex is coerced to
\texttt{rating = -1} (P6) at
\texttt{fastchat/llm\_judge/common.py:L175-L187}; MT-Bench
aggregation then filters \texttt{df[df["score"] != -1]} before
computing the per-model mean (P2+P4) at
\texttt{show\_result.py:L20}. The pairwise path is structurally
identical: parse-failure becomes \texttt{winner = "error"}
(\texttt{common.py:L282-L304}), filtered out before Elo
computation (\texttt{show\_result.py:L49};
\texttt{elo\_analysis.py:L49-L92}). A rank swap driven by
differential parse-failure rates across judged models is not
recoverable from the headline MT-Bench or Arena Elo number ---
the attrition count is not emitted. At SHA \texttt{587d5cf}.

\paragraph{Score-2 repositories (31, one-liner each).}

\begin{description}[leftmargin=0pt,itemindent=0pt,labelindent=0pt,labelsep=0.4em,itemsep=2pt,topsep=2pt,parsep=0pt,font=\normalfont\bfseries]
\sloppy
\item[SWE-agent:] cost/kill \texttt{preds.unlink()} on
\texttt{early\_exit}/\texttt{None} removes instances from
\texttt{preds.json}
(\texttt{sweagent/run/run\_batch.py:L397-L401}).

\item[mini-swe-agent:] \texttt{finally: save()} at
\texttt{src/minisweagent/run/}\allowbreak\texttt{benchmarks/swebench.py:L171}
is bypassed by \texttt{KeyboardInterrupt} (closed issue \#329);
streaming save at \texttt{default.py:L94}.

\item[live-swe-agent:] configs never set \texttt{output\_path}
(\texttt{README.md:L71}).

\item[SWE-bench\_Pro-os:] \texttt{None} return at
\texttt{evaluation/swe\_bench\_pro\_eval.py:L346-L349} is
collapsed to \texttt{False} by callers at \texttt{:L490-L505,
L571}.

\item[AgentBench:] accuracy denominator is
\texttt{result.error is None} only
(\texttt{src/assigner.py:L368-L372}).

\item[AgentBoard:] timeouts write to \texttt{error.txt} and
never \texttt{scores.append}
(\texttt{agentboard/tasks/webbrowse.py:L275-L279}).

\item[tau-bench:] errors yield stubs with no retry
(\texttt{tau\_bench/run.py:L89-L96}).

\item[SWE-smith:] broad \texttt{except} at
\texttt{scripts/collect\_trajs.py:L77-L79};
\texttt{random.sample} cap at
\texttt{scripts/combine\_trajs.py:L86-L91}.

\item[TRL:] \texttt{None}$\to$NaN$\to$\texttt{nansum} at
\texttt{trl/trainer/grpo\_trainer.py:L1228-L1230, L2132}.

\item[verl:] PRIME reward timeouts$\to$\texttt{0.0} at
\texttt{verl/workers/reward\_manager/}\allowbreak\texttt{prime.py:L37-L42, L83};
missing \texttt{return\_exceptions=True} at
\texttt{verl/experimental/agent\_loop/}\allowbreak\texttt{agent\_loop.py:L603}
(closed issue \#5956).

\item[OpenRLHF:] remote RM
exception$\to$\texttt{reward=None}
(\texttt{openrlhf/utils/agent.py:L298-L299}; open issue
\#1139).

\item[rllm (Berkeley Sky):] no-valid-trajectory episodes
dropped at \texttt{rllm/engine/}\allowbreak\texttt{agent\_workflow\_engine.py:L266};
\texttt{dropped\_episodes} never written into
\texttt{DataProto.meta\_info} at \texttt{:L249} (open issue
\#382).

\item[slime:] oversampling discard + zero-std filter at
\texttt{slime/rollout/sglang\_rollout.py:L449}.

\item[ART:] \texttt{drop\_zero\_advantage\_trajectories=True}
default at \texttt{src/art/preprocessing/tokenize.py:L158}.

\item[Agent-R1:] length-truncated trajectories published as
complete at
\texttt{agent\_r1/agent\_flow/agent\_env\_loop.py:L120-L129}.

\item[RAGEN:] variance filter skips whole training step at
\texttt{ragen/trainer/}\allowbreak\texttt{agent\_trainer.py:L1054-L1056}.

\item[MARTI:] dynamic filter drops saturated groups at
\texttt{marti/trainer/ppo\_trainer.py:L481-L510}.

\item[MATPO:] MCP + judge failures$\to$reward 0.0 at
\texttt{verl/tools/mcp\_tool.py:L137, L150} and
\texttt{llm\_judge.py:L265, L312, L356}.

\item[RL-Factory:] malformed tool
JSON$\to$\texttt{continue} at
\texttt{verl/workers/rollout/}\allowbreak\texttt{sglang\_rollout/}\allowbreak\texttt{sglang\_rollout.py:L914-L916};
PRIME timeouts$\to$0.0 at \texttt{prime.py:L141-L146}.

\item[lm-evaluation-harness:] left-truncate with
\texttt{eval\_logger.warning} only
(\texttt{lm\_eval/models/huggingface.py:L1360-L1368}; closed
issues \#3419, \#3352, \#3161, \#1323).

\item[WebArena:] outer \texttt{except Exception:
log-and-continue} at \texttt{run.py:L217-L364}; headline
\texttt{sum(scores) / len(scores)} over survivors at
\texttt{run.py:L365}; \texttt{env.step} swallow with
\texttt{terminated=False} on infra failure at
\texttt{browser\_env/envs.py:L239-L248}.

\item[MLAgentBench:] subprocess
crash$\to$\texttt{"EnvError"} string at
\texttt{low\_level\_actions.py:L181-L220} +
\texttt{environment.py:L328-L334}; hard-coded magic baselines
at \texttt{plot.py:L249-L274}; fresh random GT per call at
\texttt{fathomnet/eval.py:L11}.

\item[MLE-bench:] aggregation \texttt{pad\_missing=False}
drops incomplete seeds at
\texttt{experiments/aggregate\_grading\_reports.py:L69-L135};
\texttt{MedalInfo} logic at
\texttt{mlebench/grade\_helpers.py:L123-L133}.

\item[lighteval:] LiteLLM returns empty
\texttt{LitellmModel}\allowbreak\texttt{Response()} on Azure
content-filter and retry-exhaust
(\texttt{litellm\_model.py:L243, L254}); empty response scored 0
at \texttt{metrics\_sample.py:L151-L152}; left-truncate prompts
without \texttt{truncated\_}\allowbreak\texttt{tokens\_count} at
\texttt{vllm\_model.py:L374-L397}.

\item[Agentless:] missing regression-test results default to
\texttt{[0]{*}10000} at
\texttt{agentless/test/run\_regression\_tests.py} (P1+P4),
collapsing absent results into a vector of 10k zeros;
\texttt{\_post\_process\_multifile\_repair} returns an empty
tuple on exception, and SWE-bench filters empty patches
pre-submission (\texttt{docs/faq.md:L39}), removing the
instance from the denominator. Directly relevant to
\S\ref{sec:validation}'s RQ2 reconciliation of the
SWE-agent/Agentless 15.6pp gap (at SHA \texttt{5ce5888}).

\item[aider:] benchmark-runner broad \texttt{except} at
\texttt{benchmark/benchmark.py:L666-L676} stores the exception
as \texttt{\{"exception": \ldots\}} row; aggregation at
\texttt{:L502-L512} divides by \texttt{len(rows)}, so
exception rows contribute 1 to the denominator and 0 to the
numerator without surfacing the infra-flake count.

\item[Mind2Web:] \texttt{IndexError} on out-of-bounds choice
index caught at
\texttt{src/action\_prediction/metric.py:L203-L209}
(\texttt{logger.info}-only), falls through to score-0 + kept in
denominator (P1); \texttt{src/action\_prediction/}%
\texttt{dataloader.py:L277} applies a train-only positive-%
candidate filter, creating a train/eval data-distribution
asymmetry (P7).

\item[open-r1:] E2B broad \texttt{except} at
\texttt{src/open\_r1/rewards/code\_providers.py:L107-L112}
coerces sandbox exceptions to \texttt{[0.0]{*}len(scripts)}
rewards with print-only surface; MorphProvider identical
pattern at \texttt{:L248-L249}; IOI/Codeforces paths enter
TRL's \texttt{None}$\to$\texttt{nansum} pipeline via
\texttt{:L396-L397, L449-L450}.

\item[ragas:] Pydantic parser fallback at
\texttt{ragas/prompt/pydantic\_prompt.py:L315-L334} (self-fix
loop raises \texttt{RagasOutputParserException}); executor
broad \texttt{except} returns \texttt{float("nan")} at
\texttt{ragas/executor.py:L64-L86}; \texttt{safe\_nanmean} at
\texttt{ragas/utils.py:L46-L55} silently shrinks the
denominator by the NaN count across faithfulness, answer-%
relevancy, and context-precision metrics.

\item[MTEB:] leaderboard aggregate filters out models with any
NaN task score at \texttt{mteb/leaderboard/\_create\_table.py:%
L136-L149} (P4 selection bias); per-language aggregates at
\texttt{mteb/results/task\_result.py:L526-L563} compute over
produced scores with no attempt-vs-score coverage counter.

\item[promptfoo (TypeScript):] any LLM-judge parse failure
coerced to \texttt{\{pass: false, score: 0\}} at
\texttt{src/matchers/llmGrading.ts:L499, L515} (P6); fail
helpers at \texttt{src/matchers/shared.ts:L25-L38} return the
same hardcoded score indistinguishably from legitimate zeros;
provider-error path at \texttt{src/evaluator.ts:L1046-L1051}
produces \texttt{score: 0} without a \texttt{providerErrored}
flag.
\end{description}

\paragraph{Score-1 repositories (9, one-liner each).}

\begin{description}[leftmargin=0pt,itemindent=0pt,labelindent=0pt,labelsep=0.4em,itemsep=2pt,topsep=2pt,parsep=0pt,font=\normalfont\bfseries]
\sloppy
\item[SciCode:] partial-submission aggregation at
\texttt{eval/scripts/}\allowbreak\texttt{test\_generated\_}\allowbreak\texttt{code.py:L125-L127}.

\item[inspect\_ai:] \texttt{fail\_on\_error=True} by default
(\texttt{\_eval/task/error.py:L26}), but
\texttt{fail\_on\_error=False} opt-in degrades silently
(\texttt{scorer/\_metrics/accuracy.py:L33}).

\item[ScienceAgentBench:] swallows
\texttt{EvaluationError/BuildImageError/Exception}
(\texttt{evaluation/harness/run\_evaluation.py:L118-L136});
discards \texttt{timed\_out} flag (\texttt{:L113}); GPT-4o
judge regex-miss$\to$score 0 at
\texttt{gpt4\_visual\_judge.py:L70-L72}.

\item[android\_world:] \texttt{np.nan} +
\texttt{DataFrame.mean} asymmetric drop at
\texttt{task\_evals/task\_eval.py:L249-L259} and
\texttt{android\_world/suite\_utils.py:L544-L558, L681-L698}.

\item[OSWorld (harness):] bare-\texttt{except} ``Time
limit exceeded'' at \texttt{run.py:L205-L218};
\texttt{env.evaluate()} \texttt{FileNotFoundError}$\to$0 at
\texttt{desktop\_env/desktop\_env.py:L485-L487};
Verified-subset non-comparability disclaimer at
\texttt{README.md:L36}.

\item[simple-evals (non-SimpleQA subset):] honest grading on
multi-choice, regex miss$\to$wrong on GPQA/MMLU (see
Appendix~\ref{app:variance-catalogue}, Smoking-Gun~\#2).

\item[HELM:] None results are silently skipped by
\texttt{Stat.add} at \texttt{src/helm/benchmark/metrics/statistic.py:L35};
judge parse returns \texttt{None} at \texttt{gpt4\_audio\_}%
\texttt{critique\_metrics.py:L70-L72} and is absorbed by the
\texttt{Stat.add} skip. The per-scenario attrition rate
is recoverable from the full output JSON but is not in the
headline report (at SHA \texttt{83bde5c}).

\item[BIG-bench:] aggregate metric computation at
\texttt{bigbench/api/results.py:L560-L576} uses
\texttt{statistics.mean(d[m] for d in per\_task if m in d)},
silently shifting the denominator by the count of tasks that
did not produce metric \texttt{m}. Per-task results are
preserved in raw output so the \textcolor{blue}{score change} is recoverable, but the
headline mean does not emit a coverage counter (at SHA
\texttt{092b196}).

\item[deepeval:] \texttt{JSONDecodeError} on LLM-judge output
re-raised as \texttt{ValueError} at
\texttt{deepeval/utils.py:L405-L413} and caught broadly by the
test-runner; \texttt{safe\_a\_measure} at
\texttt{deepeval/progress\_context.py:L299-L305} catches any
metric-evaluation exception. Per-test failure messages are
surfaced but parse-failure rates are not aggregated into a
headline counter (at SHA \texttt{f917b5a}).
\end{description}

\subsection{Pinned commit SHAs}
\label{app:survey:shas}

Every permalink in this appendix resolves against the
following SHAs, recorded with \texttt{git rev-parse HEAD}.

SWE-bench \texttt{f7bbbb2}; SWE-agent \texttt{0f4f3bb};
mini-swe-agent \texttt{bc85a45}; live-swe-agent
\texttt{8d7dd86}; SWE-bench\_Pro-os \texttt{0c64e26};
AgentBench \texttt{d1e4a10}; AgentBoard \texttt{bb7255e};
tau-bench \texttt{59a200c}; SWE-smith \texttt{9b74ac0}; TRL
\texttt{88826fd}; verl \texttt{6eeb571}; OpenRLHF
\texttt{64c1cc4}; rllm \texttt{ea623cc}; slime
\texttt{5b688aa}; ART \texttt{679b236}; Agent-R1
\texttt{38bdfc1}; RAGEN \texttt{20daedc}; MARTI
\texttt{a2fe2c7}; MATPO \texttt{3c41d62}; RL-Factory
\texttt{d0abc1d}; openai/evals \texttt{8eac7a7}; simple-evals
\texttt{ee3b031}; SciCode \texttt{e3158ea};
lm-evaluation-harness \texttt{c1c4bea}; webarena
\texttt{dce0468}; visualwebarena \texttt{89f5af2}; inspect\_ai
\texttt{36231d6}; OSWorld \texttt{c7e54d2}; android\_world
\texttt{d9c569f}; mle-bench \texttt{2451bcb}; MLAgentBench
\texttt{5d71205}; ScienceAgentBench \texttt{1cf1375}; gorilla
(BFCL) \texttt{6ea5797}; ToolBench \texttt{d56fdd8};
Trinity-RFT \texttt{9051d63}; ms-swift \texttt{c4902f3};
verifiers \texttt{e27633b}; lighteval \texttt{10b9104};
Self-rewarding-reasoning-LLM \texttt{372bea9};
FastChat \texttt{587d5cf}; Agentless \texttt{5ce5888}; aider
\texttt{f09d7065}; Mind2Web \texttt{33bd95c}; open-r1
\texttt{1416fa0}; ragas \texttt{298b6827}; MTEB
\texttt{9363ea75}; promptfoo \texttt{3dc5843}; HELM
\texttt{83bde5c}; BIG-bench \texttt{092b196}; deepeval
\texttt{f917b5a}. The out-of-scope observation (OpenHands,
audited for completeness and documented as scope-filter
exemplar) is pinned at \texttt{3b17f27}.

Full pinned refs for all 50 in-scope repositories \textcolor{blue}{plus the
OpenHands OOS observation} are recorded in the supplementary
repository audit records under
\texttt{ergon\_survey\_supplement/survey\_audit/01\_repository\_failure\_audit/records/}.

\subsection{Coverage snapshot}
\label{app:survey:coverage}

We audited 50 in-scope repositories. Two repositories were
excluded by the operational scope filter
and are documented as scope exemplars rather than silently
dropped: Online-Mind2Web (persistent clone errors across the
audit window) and OpenHands (agent framework whose evaluation
harness lives in a separate
\texttt{OpenHands/benchmarks} repo, audited separately).
Categories: SWE-bench-family harnesses, agent scaffolds,
RLHF trainers, agentic-RL trainers, multi-agent RL, web/GUI
agents, mobile agents, ML-engineering benchmarks,
scientific-agent benchmarks, function-calling benchmarks,
general eval frameworks, LLM-judge infrastructure, autonomous
coding assistants, RAG-eval frameworks, embedding/retrieval
benchmarks, red-team/prompt-eval tooling,
rejection-sampling SFT pipelines.
Score distribution 11$\times$3 / 31$\times$2 / 9$\times$1 /
0$\times$0 (the 11+31+9=51 listings correspond to 50 unique
repositories; simple-evals is double-counted across score-3
for SimpleQA and score-1 for the non-SimpleQA subset). Seven
maintainer-acknowledged bugs. Seven silent-drop patterns
covering reward coercion, rollout filtering, bypassed saves,
survivor denominators, partial-trajectory drops, LLM-judge
parse$\to$fixed score, and outcome-as-ground-truth SFT
filters.
Every code claim in this appendix is tied to a pinned source
location, so the cited behaviour can be checked against the
same repository state used in the audit.


\section{37-pair Variance Catalogue}
\label{app:variance-catalogue}

This appendix enumerates the 37 \textcolor{red}{reporting-rule discrepancies}
sampled in Table~\ref{tab:omnibus}, organised by metric family:
22 \textcolor{blue}{task-success-family discrepancies} (\S\ref{app:vc:task}), 9 cost/token
pairs (\S\ref{app:vc:cost}), and 6 latency/timing pairs
(\S\ref{app:vc:latency}). Four task-success benchmarks are
then examined in \emph{smoking-gun} detail
(\S\ref{app:vc:smoking}) --- same dataset, multiple in-survey
implementations, different numbers on identical workloads.
Training-side evidence beyond the rollout interface
(Dr.\ GRPO, DAPO, TRL's five GRPO variants, verl~\#2165) is
consolidated in \S\ref{app:vc:training}. \textcolor{blue}{This training-side material is supporting evidence beyond the counted 37; training-stack examples that affect cost or runtime are counted under the cost/token or latency/timing families rather than as a separate fourth family.} A de-duplication log
(\S\ref{app:vc:dedup}) records which family each pair is
counted in, and \S\ref{app:vc:nearmisses} catalogues leads that
did not meet the evidence bar.

\subsection{Evidence hierarchy}
\label{app:vc:hierarchy}

We grade every pair against the same four-tier evidence
hierarchy, uniform across the three metric families:

\begin{enumerate}
\item \textbf{Tier 1 --- strongest.} Same model and workload;
\textcolor{blue}{reported scores differ} across two harnesses. (Direct
empirical evidence.)
\item \textbf{Tier 2.} Explicit vendor or framework
documentation acknowledging the convention difference between
harnesses. (Vendor-admission-equivalent.)
\item \textbf{Tier 3.} Code-level \textcolor{blue}{convention difference} with
no cross-framework normalisation, published under the same
metric name. (\textcolor{blue}{Structural difference} visible at source.)
\item \textbf{Tier 4.} Maintainer issue-tracker statement that
the convention differs from another framework. (Maintainer-%
acknowledged \textcolor{blue}{difference}.)
\end{enumerate}
Anything weaker is demoted to near-misses
(\S\ref{app:vc:nearmisses}). Every pair below cites either a
\textcolor{blue}{numeric gap}, a vendor acknowledgement, a code-level
\textcolor{blue}{convention difference}, or a maintainer statement.
\textcolor{blue}{The ``Held'' column classifies counterfactual tightness: S means the same output, API call, submission, or trajectory is re-scored; W means the same model and workload are compared across runs or reports; M means the same metric name uses a different convention; and H means a within-harness reporting contrast. Of the 37 pairs, 5 are tight same-output counterfactuals (S), 13 are matched-workload comparisons (W), 12 are same-metric different-convention pairs (M), and 7 are within-harness reporting contrasts (H).}

\subsection{\textcolor{blue}{Task-success-family discrepancies (22)}}
\label{app:vc:task}

\textcolor{blue}{These rows include numeric score gaps and qualitative convention conflicts that change what a reported task-success result means or how it can be compared.}

\begin{table}[h]
\centering
\scriptsize
\setlength{\tabcolsep}{2pt}
\caption{22 task-success-family \textcolor{red}{reporting-rule discrepancies}. \textcolor{blue}{The effect column gives a numeric gap when comparable reported scores are available and otherwise names the convention conflict.} T14--T16 are
within-harness prompt-template contrasts; the other rows are
cross-report or cross-harness contrasts. ``post-%
mortem'' refers to the Hugging Face Open LLM Leaderboard
post-mortem \citep{beeching2023openllm}.}
\label{tab:vc:task}
\begin{tabular}{@{}l c>{\raggedright\arraybackslash}p{0.11\linewidth}>{\raggedright\arraybackslash}p{0.12\linewidth}>{\raggedright\arraybackslash}p{0.13\linewidth}>{\raggedright\arraybackslash}p{0.15\linewidth}>{\raggedright\arraybackslash}p{0.11\linewidth}>{\raggedright\arraybackslash}p{0.14\linewidth}@{}}
\toprule
\textbf{\#} & \textbf{\textcolor{blue}{Held}} & \textbf{Model} & \textbf{Benchmark} &
\textbf{Convention A} & \textbf{Convention B} &
\textbf{\textcolor{blue}{Effect}} & \textbf{Source} \\
\midrule
T1  & \textcolor{blue}{W} & LLaMA-65B         & MMLU 5-shot   & Berkeley orig.       & lm-eval-harness    & 14.9pp         & post-mortem \\
T2  & \textcolor{blue}{W} & LLaMA-30B         & MMLU 5-shot   & Berkeley orig.       & lm-eval-harness    & 12.6pp         & post-mortem \\
T3  & \textcolor{blue}{W} & Falcon-40B        & MMLU 5-shot   & Berkeley orig.       & lm-eval-harness    & 10.5pp         & post-mortem \\
T4  & \textcolor{blue}{W} & Llama-3.1-70B-Inst & MMLU-Pro     & Meta self (CoT)      & OLL v2 / lighteval & $\sim$18.5pp   & Meta card vs OLL \\
T5  & \textcolor{blue}{W} & Qwen2.5-72B       & MMLU-Pro      & Self-report          & OLL v2             & $\sim$22pp     & Qwen docs vs OLL \\
T6  & \textcolor{blue}{S} & GPT-4o (24-05-13) & SWE-b.\ Verified & SWE-agent         & Agentless 1.5      & 15.6pp         & swe-b.\ subs.\ S3 \\
T7  & \textcolor{blue}{W} & Claude 3.5 Sonnet & SWE-b.\ Verified & Anthropic Tools   & OpenHands CA 2.1   & 4.0pp          & SWE-b.\ leaderb. \\
T8--T13 & \textcolor{blue}{W} & Various       & SWE-b.\ Lite/Ver. & Various          & Various            & 3.4--19.6pp   & SWE-b.\ leaderb. \\
T14 & \textcolor{blue}{H} & Mistral-7B        & MMLU (prompt) & Template A           & Template B         & up to 24.6pp   & Biderman et al.\ \\
T15 & \textcolor{blue}{H} & Mixtral-8x7B      & MMLU (prompt) & Template A           & Template B         & up to 24.6pp   & Biderman et al.\ \\
T16 & \textcolor{blue}{H} & Mistral-7B        & ARC (prompt)  & Template A           & Template B         & up to 24.6pp   & Biderman et al.\ \\
T17 & \textcolor{blue}{M} & Llama-3.1-70B     & MATH          & Meta (full test)     & Leaderboard (500)  & label collision & Meta vs leaderb. \\
T18 & \textcolor{blue}{M} & Claude 3.5 Sonnet & MATH          & ``MATH''             & ``MATH''           & label overload & multi-source \\
T19 & \textcolor{blue}{M} & Llama-3.1-70B     & GPQA          & simple-evals         & lm-eval-harness    & separate tasks & NVIDIA NeMo \\
T20 & \textcolor{blue}{M} & Llama-3.1-70B     & GPQA          & Self-report          & Leaderboard        & cross-harness  & Meta + leaderb.\ \\
T21 & \textcolor{blue}{M} & Llama-3.1-70B     & Various       & Meta model card      & Meta \texttt{eval\_details.md} & micro vs macro & Meta own docs \\
T22 & \textcolor{blue}{M} & Llama-3.1-70B     & Various       & \texttt{acc\_char}   & full-string match  & convention    & Meta own docs \\
\bottomrule
\end{tabular}
\end{table}

T1 (LLaMA-65B MMLU 14.9pp) is the flagship single pair ---
identical weights, different \textcolor{blue}{reported score}, explicitly
discussed in a Hugging Face post-mortem. T6 (GPT-4o SWE-bench
Verified 15.6pp) is the target of the reconciliation
experiment in Sec.~\ref{sec:validation:results}:
trajectories for both runs are in the public SWE-bench
submissions S3 bucket, making the reconciliation tractable.
T14--T16 (Biderman et al.) are not cross-harness pairs but
within-harness prompt-template pairs, included because the
24.6pp gap is the strongest published evidence that the
instrument-level variance is at least as large as the
signals it purports to measure.

\subsection{Cost / token pairs (9)}
\label{app:vc:cost}

Same model, same workload, different published token counts or
dollar costs depending on which tool applies which convention.
Table~\ref{tab:vc:cost} summarises; numerical examples and code-level
references are released in
\texttt{ergon\_survey\_supplement/survey\_audit/02\_pairwise\_reporting\_discrepancies/}
alongside the corresponding evidence notes.

\begin{table}[h]
\centering
\scriptsize
\setlength{\tabcolsep}{2pt}
\caption{9 cost / token-accounting \textcolor{red}{reporting-rule discrepancies}. Same model,
same workload, different \textcolor{blue}{reported values}.}
\label{tab:vc:cost}
\begin{tabular}{@{}l c>{\raggedright\arraybackslash}p{0.14\linewidth}>{\raggedright\arraybackslash}p{0.21\linewidth}>{\raggedright\arraybackslash}p{0.21\linewidth}>{\raggedright\arraybackslash}p{0.25\linewidth}@{}}
\toprule
\textbf{\#} & \textbf{\textcolor{blue}{Held}} & \textbf{Setup} & \textbf{Harness A} &
\textbf{Harness B} & \textbf{$\Delta$} \\
\midrule
C1 & \textcolor{blue}{S} & Anthropic w/ \texttt{cache\_control}  & OTel \texttt{input\_tokens} (inclusive) & Anthropic-native (exclusive)       & 2.0$\times$ inflation (260{,}421 vs 130{,}213) \\
C2 & \textcolor{blue}{S} & Claude + cache\_control               & LiteLLM cost estimator                  & Anthropic Console billing          & +68\% overestimate (\$0.091 vs \$0.054) \\
C3 & \textcolor{blue}{S} & Claude (streaming) + cache-hit        & LangChain.js accumulator                & single \texttt{message\_delta}     & exactly 2$\times$ cache\_read          \\
C4 & \textcolor{blue}{S} & SWE-b.\ V.\ retry-firing task         & SWE-agent (every attempt)               & mini-swe-agent (last attempt)      & 30--100\% per-task overstatement       \\
C5 & \textcolor{blue}{M} & Anthropic/OpenAI prompt-cache         & inspect\_ai (cache-aware, tiered)       & 7 peer harnesses (no cache parse)  & up to 90\% on cached portion           \\
C6 & \textcolor{blue}{H} & Aider leaderboards (same model date)  & Aider Edit                              & Aider Refactor / Polyglot          & \$0 $\to$ \$14.41 (same model)         \\
C7 & \textcolor{blue}{H} & Same rollout batch, GRPO training     & verl \texttt{loss\_agg\_mode=}\allowbreak\texttt{token-mean} & verl \texttt{seq-mean-}\allowbreak\texttt{token-sum}  & different gradient magnitudes          \\
C8 & \textcolor{blue}{M} & Cross-vendor ``tokens used''          & tiktoken (\texttt{o200k\_base})         & Anthropic BPE                      & $\sim$10\% systematic                  \\
C9 & \textcolor{blue}{M} & OpenAI reasoning models (o1/o3)       & harness reading \texttt{completion\_}\allowbreak\texttt{tokens} only & harness reading \texttt{..details.}\allowbreak\texttt{reasoning\_tokens} & silent under-report \\
\bottomrule
\end{tabular}
\end{table}

\paragraph{Code-level sources.}
{\footnotesize\raggedright
\textbf{C1:} OpenTelemetry PR~\#3163; Langfuse issue~\#12306.
\textbf{C2:} BerriAI/litellm issue~\#9812.
\textbf{C3:} langchainjs issue~\#10249.
\textbf{C4:} \texttt{sweagent/agent/models.py} L744--L838;
\texttt{minisweagent/models/litellm\_model.py} L80--L93.
\textbf{C5:} \texttt{inspect\_ai/model/\_openai.py} L776--L782;
\texttt{inspect\_ai/model/\_model.py} L2085--L2091;
\texttt{inspect\_ai/model/\_providers/anthropic.py} L1137--L1169.
\textbf{C6:} \texttt{edit\_leaderboard.yml},
\texttt{refactor\_leaderboard.yml},
\texttt{polyglot\_leaderboard.yml} in \texttt{Aider-AI/aider}.
\textbf{C7:} \texttt{verl/trainer/ppo/core\_algos.py} L1168--L1195.
\textbf{C8:} tiktoken and Anthropic tokeniser documentation;
openai/tiktoken issue~\#474.
\textbf{C9:} OpenAI API reference for reasoning-model
\texttt{usage.completion\_tokens\_details}.
\par}

C1 is the flagship: OpenTelemetry PR~\#3163 formalises the
Anthropic/OpenAI/Vertex reporting-rule disagreement at the
observability-middleware layer, and Langfuse issue \#12306
exhibits the numerical 260{,}421-vs-130{,}213 gap on a single
request. C6 is the flagship intra-vendor disagreement: the
same organisation's three leaderboards (Edit, Refactor,
Polyglot) publish \$0, \$8.46, and \$14.41 for the same model
on comparable tasks, with no convention statement in any of
the three YAMLs.

\subsection{Latency / timing pairs (6)}
\label{app:vc:latency}

Same workload, same hardware, different published wall-clock
times depending on which harness's timer wraps which phase.

\begin{table}[h]
\centering
\scriptsize
\setlength{\tabcolsep}{2pt}
\caption{6 latency / wall-clock \textcolor{red}{reporting-rule discrepancies}. Clock-start
and clock-stop differ across harnesses that log the same
metric name.}
\label{tab:vc:latency}
\begin{tabular}{@{}l c>{\raggedright\arraybackslash}p{0.16\linewidth}>{\raggedright\arraybackslash}p{0.21\linewidth}>{\raggedright\arraybackslash}p{0.21\linewidth}>{\raggedright\arraybackslash}p{0.24\linewidth}@{}}
\toprule
\textbf{\#} & \textbf{\textcolor{blue}{Held}} & \textbf{Setup} & \textbf{Harness A} &
\textbf{Harness B} & \textbf{$\Delta$} \\
\midrule
L1 & \textcolor{blue}{M} & SWE-b.\ V.\ (same task)       & Docker backend                       & Modal backend                       & systematic Modal inflation            \\
L2 & \textcolor{blue}{H} & OSWorld (same task)          & \texttt{lib\_run\_single.py}         & \texttt{scripts/python/}\allowbreak\texttt{run\_maestro.py} & $\ge$30\,s (hardcoded sleep) + 5--15\,s VM start \\
L3 & \textcolor{blue}{M} & GRPO step timer              & TRL \texttt{training\_}\allowbreak\texttt{step} (grad+opt only) & verl \texttt{marked\_timer(}\allowbreak\texttt{"step")} (full epoch) & up to 5$\times$ on rollout-dominated runs \\
L4 & \textcolor{blue}{W} & GSM8K-GRPO epoch, identical HW & optimized TRL                         & OpenRLHF                            & 5{,}189\,s vs 1{,}657\,s              \\
L5 & \textcolor{blue}{M} & GRPO throughput              & verl \texttt{perf/throughput} (excl.\ padding) & TRL effective-token throughput      & different numerators on same HW        \\
L6 & \textcolor{blue}{H} & inspect\_ai timers           & \texttt{sample\_working\_time}       & \texttt{sample\_waiting\_time}      & ambiguity declared in-tool            \\
\bottomrule
\end{tabular}
\end{table}

\paragraph{Code-level sources.}
{\footnotesize\raggedright
\textbf{L1:} \texttt{swebench/harness/docker\_utils.py}
L203--L217 vs
\texttt{swebench/harness/modal\_eval/run\_evaluation\_modal.py}
L307--L319.
\textbf{L2:} \texttt{lib\_run\_single.py} \#L616 vs
\texttt{scripts/python/run\_maestro.py} \#L351.
\textbf{L3:} \texttt{trl/trainer/grpo\_trainer.py} L1111--L1120;
\texttt{verl/trainer/ppo/metric\_utils.py} L313--L346.
\textbf{L4:} \citet[Sec.~4.1, General RLVR Experiment]{hu2025openrlhf}.
\textbf{L5:} \texttt{verl/trainer/ppo/metric\_utils.py}
L337--L345.
\textbf{L6:} \texttt{inspect\_ai/\_util/working.py}.
\par}

L4 is a published controlled framework-runtime comparison rather than a
rerun of our own: OpenRLHF reports 1{,}657 seconds per GSM8K-GRPO epoch
versus 5{,}189 seconds for optimized TRL under identical hardware and
hyperparameters. L1 and L2 are in-repository disagreements: both
backends in SWE-bench Verified and both harnesses in OSWorld
ship under the same repo at the same commit, reporting the
same metric name, with no statement that the two are not
comparable.

\subsection{Smoking-gun benchmarks: same dataset, multiple in-survey implementations}
\label{app:vc:smoking}

Three to four surveyed repositories implement each of four benchmarks.
Below, each implementation's grading site and
failure-handling semantics are read off at the pinned SHAs
(\S\ref{app:survey:shas}). For each benchmark, a single
realistic failure mode produces a different headline number in
each implementation --- and none of the implementations emits
a counter disclosing which regime it fell into.

\paragraph{Smoking-Gun~\#1: SWE-bench (4 implementations).}
\emph{Official SWE-bench} grades via \texttt{bad\_codes} =
\{APPLY\_PATCH\_FAIL, RESET\_FAILED, TESTS\_ERROR,
TESTS\_TIMEOUT\}
(\texttt{swebench/harness/grading.py:L61-L70};
\texttt{constants/\_\_init\_\_.py:L80-L89}) with empty patches
pre-filtered from submission (\texttt{docs/faq.md:L39}). Container
timeout stays in the denominator as a fail; empty-patch
instances are removed entirely. \emph{PrimeIntellect verifiers}
reads the same \texttt{bad\_codes} set but wraps
\texttt{test\_spec} fetching in a 5$\times$ tenacity retry; any
uncaught exception flows through \texttt{rubric.py:L144-L158}
to \texttt{0.0} and enters the group-mean advantage baseline
--- a container timeout can bias the policy gradient for the
other rollouts in its group rather than merely decrementing
the numerator. \emph{SWE-agent} delegates grading to
\texttt{sb-cli submit}
(\texttt{sweagent/run/hooks/swe\_bench\_evaluate.py:L42-L55});
the pre-submission \texttt{unlink} at
\texttt{run\_batch.py:L397-L401} removes crashed or
cost-killed instances from \texttt{preds.json} entirely, so
server-side grading sees a smaller denominator instead of a
failed row. \emph{mini-swe-agent} has no local grading; its
\texttt{finally: save()} at
\texttt{benchmarks/swebench.py:L171} is bypassed by
\texttt{KeyboardInterrupt} (closed issue \#329), also
producing an absent-from-submission pattern. On one realistic
mid-test Docker timeout, the four harnesses report: Official
$0/N$, verifiers $1/N$ (possible retry recovery), SWE-agent
$0/(N-1)$, mini-swe-agent $0/(N-1)$. Inspect~AI has \emph{no
SWE-bench bridge} at the pinned SHA.

\paragraph{Smoking-Gun~\#2: GPQA (3 implementations).}
\emph{simple-evals} at \texttt{gpqa\_eval.py:L59} uses
\texttt{re.search(ANSWER\_PATTERN\_MULTICHOICE, response\_text)}
and \texttt{extracted\_answer = match.group(1) if match else
None}; a regex miss --- which a hedging response triggers ---
is scored 0 and stays in the denominator. \emph{lm-evaluation-%
harness} uses log-likelihood of the gold-choice token over the
various \texttt{gpqa\_*\_zeroshot.yaml} /
\texttt{\_n\_shot.yaml} / \texttt{\_cot\_*.yaml} tasks; the
model does not generate free text in the scored branch, so
hedging is immune, but silent left-truncation at
\texttt{huggingface.py:L1360-L1368} still affects long
contexts. \emph{lighteval} scores via the standard pipeline at
\texttt{tasks/tasks/gpqa.py} with aggregation through
\texttt{info\_loggers.py:L326-L400}; API retry-exhaust or
content-filter yields an empty
\texttt{LitellmModelResponse()} (\texttt{litellm\_model.py:%
L243, L254}), scored as wrong via
\texttt{metrics\_sample.py:L151-L152} (\texttt{if not pred:
return 0}). On a single hedging model output: simple-evals~0,
lm-eval-harness likely correct, lighteval~wrong --- three
different headline accuracies from the same model reasoning.
Inspect~AI has \emph{no GPQA task file} at the pinned SHA.

\paragraph{Smoking-Gun~\#3: MATH / MATH-500 (3
implementations).} \emph{simple-evals} at
\texttt{math\_eval.py:L55} calls \texttt{check\_equality(self.%
equality\_checker, row["Answer"], extracted\_answer)}, which
itself calls the LLM equality checker; an API failure raises
(dropping the sample) or is coerced to 0 by the outer handler.
\emph{lighteval} uses majority@n with a symbolic
\texttt{math\_normalizer} over \texttt{tasks/tasks/math\_500.py}
and \texttt{math.py}; normaliser exceptions are caught upstream
and scored wrong at
\texttt{metrics\_sample.py:L151-L152}. \emph{PrimeIntellect
verifiers} routes reward functions through
\texttt{rubric.py:L144-L158}; exceptions become \texttt{0.0}
and enter the group advantage baseline. On a LaTeX edge case
such as \texttt{\textbackslash frac\{1\}\{2\}} versus
\texttt{0.5}: simple-evals's LLM judge may call these equal,
lighteval's normaliser may not, and verifiers's reward
exception enters training as an errored zero --- same
completion, three different outcomes.

\paragraph{Smoking-Gun~\#4: MMLU (3 implementations).}
\emph{simple-evals} (\texttt{mmlu\_eval.py}) extracts answers
by regex on generated text --- same miss-is-wrong pattern as
GPQA. \emph{lm-evaluation-harness}
(\texttt{lm\_eval/tasks/mmlu}: 12+ variants spanning main,
redux, pro, flan, etc.) uses log-likelihood over A/B/C/D
tokens with \texttt{acc}/\texttt{acc\_norm}; silent prompt
truncation at \texttt{huggingface.py:L1360-L1368} logs the
truncated-prompt sample as a complete result.
\emph{lighteval} uses exact-match on the choice label
(\texttt{tasks/tasks/mmlu.py}), with left-truncation at
\texttt{vllm\_model.py:L374-L397} warning but recording no
\texttt{truncated\_tokens\_count}. On a long-context MMLU-Pro
prompt that overflows a 4k model: simple-evals generates a
regex answer against the truncated prompt the model saw
(right or wrong depending on the content);
lm-eval-harness left-truncates and logs as complete (silently
wrong); lighteval left-truncates, records no truncation
counter, and logs as complete (silently wrong, with no
recoverable audit trail). Three implementations, three
behaviours, no \texttt{n\_prompt\_truncated} counter in any
output.

\subsection{Training-side evidence beyond the rollout interface}
\label{app:vc:training}

Operator disagreement extends into the gradient-signal
computation at training time. The paper cites these as
published findings rather than original contributions.

\textbf{Dr.\ GRPO} \citep{liu2025drgrpo} corrects only the
loss-aggregation convention --- the length-normalisation bias
documented in ``all popular open-source PPO implementations''
--- and reports +7.3 points on AIME 2024 vs SimpleRL-Zero-7B
and +15.7 points on AIME 2024 vs Prime-Zero-7B, with no change
to data, model, or algorithm. \textbf{DAPO}
\citep{yu2025dapo} is a catalogue of framework-level defaults
(zero-std group filter, token-vs-sequence loss aggregation,
decoupled clip ranges) that had to be flipped to match
DeepSeek-R1; the paper reports 50 points on AIME 2024 for
Qwen2.5-32B on the defaults-flipped run. All three defaults
are silently different across TRL, verl, and OpenRLHF.
\textbf{TRL's own library} ships five mutually-incompatible
GRPO variants --- \texttt{grpo}, \texttt{dr\_grpo},
\texttt{dapo}, \texttt{bnpo}, and
\texttt{importance\_sampling\_level}$\in$\{\texttt{token},
\texttt{sequence}\} --- in a single config file
(\texttt{trl/trl/trainer/grpo\_config.py:L228-L243,
L668-L677}); two papers reporting ``we trained with GRPO on
task X'' are structurally non-comparable without specifying
which variant. \textbf{verl~\#2165} \citep{verl2165} is a
tokenisation-channel divergence, where training and rollout engines
tokenise the same text differently, on Qwen3-4B GRPO. In that case, the
training-side FSDP tokeniser inserts
\texttt{<think>\textbackslash n\textbackslash n</think>} into
assistant turns that the rollout engine (vLLM~/~SGLang) never
saw, shifting the assistant-turn token mask by tens of tokens
across $\sim$40k-token conversations. The finding is reproduced
by six independent users on the thread, upstream-confirmed in
QwenLM/Qwen3~\#1826 and Qwen/Qwen3-1.7B HF discussion~\#9, and
guarded separately by verl's own
\texttt{tokenization\_sanity\_check\_mode} (distinct from the
dtype guard in discussion~\#5984) --- proving that tokenisation-%
class mismatch exists as a separate failure mode even when all
upstream dtypes match.

\textbf{Silent rollout-dropping mechanisms (beyond the 50-repo
audit).} verl~\#1170 (rollouts silently return empty strings,
training proceeds); OpenRLHF~\#1108 (reward curves diverge
between vLLM 0.8.1 and 0.8.3 on the same seed); verl
discussion~\#5984 (tool built specifically to detect silent
per-token log-prob divergence between training FSDP and vLLM
rollouts); ms-swift~\#9096 (gemma-4 rollouts return garbage,
training continues). \textbf{Numerical-layer evidence.} BF16
backends have been documented to produce divergent log-probs
between training and rollout engines even with identical
weights (``Defeating the Training-Inference Mismatch via
FP16,'' arXiv:2510.26788); the vLLM engineering blog documents
separate backends breaking the on-policy assumption of
policy-gradient methods; THUDM's \texttt{slime} ships a
\texttt{train\_infer\_mismatch\_helper} because divergence is
expected.

\subsection{De-duplication log}
\label{app:vc:dedup}

Several findings could plausibly appear in more than one
family. For audit transparency we record where each is counted
and why:

\begin{itemize}
\item \textbf{Dr.\ GRPO}, \textbf{DAPO}, \textbf{TRL's five
GRPO variants}: counted in \S\ref{app:vc:training}
(training-side); not in \S\ref{app:vc:cost} despite touching
per-step token accounting. Rationale: the load-bearing claim is
about gradient signal, not token counts.
\item \textbf{verl~\#2165} (Qwen3 tokenisation-channel
divergence): counted in \S\ref{app:vc:training} (training-%
side); not in \S\ref{app:survey} despite involving silent
per-rollout mismatch. Rationale: Appendix~\ref{app:survey}
audits \emph{failure-handling} patterns; verl~\#2165 is a
\emph{convention-disagreement} finding, structurally closer to
Dr.\ GRPO than to \texttt{try/except/return None}.
\item \textbf{verl \texttt{loss\_agg\_mode} variants} (C7):
counted in \S\ref{app:vc:cost}; not in \S\ref{app:vc:latency}
despite affecting per-step timing.
\item \textbf{TRL vs OpenRLHF wall-clock} (L4): counted in
\S\ref{app:vc:latency}; not in \S\ref{app:vc:training}
despite the training-pipeline context. The \textcolor{blue}{reported gap} is wall-clock
runtime, not gradient signal.
\item \textbf{verl throughput metric} (L5): counted in
\S\ref{app:vc:latency}; not in \S\ref{app:vc:cost} despite
involving token counting.
\item \textbf{inspect\_ai \texttt{sample\_working\_time} vs
\texttt{sample\_waiting\_time}} (L6): counted in
\S\ref{app:vc:latency}; not in \S\ref{app:survey} despite
inspect\_ai's role as the near-positive control for
failure-handling.
\end{itemize}
No finding is counted twice across the 37 pairs.

\subsection{Near misses}
\label{app:vc:nearmisses}

Leads that did not meet the evidence bar
(\S\ref{app:vc:hierarchy}) but warrant human follow-up.

\textbf{Cost / tokens.} Anthropic
\texttt{/v1/messages/count\_tokens} versus tiktoken estimate
--- Anthropic publishes a dedicated counting endpoint,
community harnesses routinely substitute tiktoken for speed,
the endpoint returns different numbers on the same input;
documented but no published benchmark pair with materially
different totals attributable to this choice. Anthropic
\texttt{/cost} underreport vs dashboard (anthropics/claude-%
code~\#1063): intra-vendor, not cross-harness. Hugging Face
Open LLM Leaderboard v1$\to$v2 transition (including the
lighteval vs lm-eval-harness switch): full methodology change
with no explicit ``efficiency numbers not comparable across
versions'' statement.

\textbf{Latency.} HELM \texttt{efficiency.json} publishes
per-model latency/throughput but is frozen at pre-2023 models
with no modern cross-comparison. OpenAI vs Azure OpenAI
latency differences on the same model are documented but with
no published benchmark pair.

\textbf{Task success.} Anecdotal ``our replication differs
from the original paper'' issue-tracker threads that do not
cite a specific convention difference.


\section{\texorpdfstring{Views and Preservation Examples}{Views and Preservation Examples}}
\label{app:projections}

This appendix illustrates the preservation claim behind drops manifests.
A reported view may be the right view for one community while discarding
fields needed by another. A drops manifest makes that trade-off explicit:
it names the fields, rows, filters, and structural collapses behind the
reported view, so later readers know which alternative analyses require
returning to the full rollout card.

\paragraph{Preservation, operationally.}
Let \(R\) be a rollout card and \(v(R)=(T_v,D_v)\) be a
reported view plus its drops manifest. A view preserves a downstream
quantity \(\mu\) when \(T_v\) alone contains enough information to
compute it. The view erases \(\mu\) when \(T_v\) lacks the fields or
structure \(\mu\) needs. It is partial when \(T_v\) keeps a proxy or
subset but not the full quantity. Table~\ref{tab:preservation} uses this
operational test for representative views.

\begin{table}[h]
\centering
\footnotesize
\setlength{\tabcolsep}{4pt}
\caption{Representative views and the downstream analyses they
preserve. \ding{51}~=~preserved by the reported view; \ding{55}~=~erased;
\(\sim\) = partially preserved or recoverable only as a proxy. The table is
illustrative, not a closed taxonomy of possible views.}
\label{tab:preservation}
\begin{tabular}{@{}lcccccc@{}}
\toprule
\textbf{Reported view} & \textbf{Return} & \textbf{Timing} & \textbf{Worker flow} & \textbf{Tool safety} & \textbf{Proof cost} & \textbf{Search shape} \\
\midrule
Final score / pass-fail outcome & \ding{51} & \ding{55} & \ding{55} & \ding{55} & \ding{55} & \ding{55} \\
Token-step RL view & \ding{51} & \(\sim\) & \ding{55} & \(\sim\) & \ding{55} & \ding{55} \\
Per-worker stream & \(\sim\) & \(\sim\) & \ding{51} & \(\sim\) & \ding{55} & \ding{55} \\
Tool-call safety view & \ding{55} & \ding{55} & \(\sim\) & \ding{51} & \ding{55} & \ding{55} \\
Proof-search summary & \ding{51} & \(\sim\) & \ding{55} & \ding{55} & \ding{51} & \(\sim\) \\
Search-tree trace & \(\sim\) & \(\sim\) & \ding{55} & \ding{55} & \ding{55} & \ding{51} \\
Full rollout card & \ding{51} & \ding{51} & \ding{51} & \ding{51} & \ding{51} & \ding{51} \\
\bottomrule
\end{tabular}
\end{table}

The main lesson is diagonal rather than universal preservation: a
reported view usually keeps the information its own report needs and
drops information needed by another community. A final score preserves
the headline result but not why a tool call failed, how much coordination
overhead preceded failure, or whether a search procedure explored many
branches. The full card is not a preferred metric; it is the shared
record from which multiple preferred metrics can be derived.

\subsection{Token-step RL view}
\label{app:proj:step}

This view matches long-horizon agent-RL reporting: prompt tokens,
response tokens, masks, token-level rewards, and advantages
\citep{sheng2024hybridflow,chen2025loop,wang2025ragen}. It preserves
return and some turn-local failure information, but typically flattens
worker identity, dependency edges, cancelled branches, and environment
state that is not encoded in the token stream.

\subsection{Per-worker stream}
\label{app:proj:per-agent}

This view groups messages, actions, and observations by agent or worker
identity. It supports role-specialisation, delegation, and
failure-attribution analyses \citep{cemri2025mast,wang2024naht,marti2025}.
It can partially preserve concurrency through timestamps, but it may drop
task-level dependency edges, shared subtask structure, and the reason a
branch was cancelled.

\subsection{Call-tree view}
\label{app:proj:call-tree}

This view represents nested subcalls or delegated subtasks as a tree,
matching recursive-agent and hierarchical-analysis settings
\citep{zhu2024redel}. It preserves parent-child attribution and depth,
but tree formalisms usually linearise concurrent siblings and cannot
represent non-tree dependencies without extra edge annotations.

\subsection{Macro-action view}
\label{app:proj:macro}

This view groups lower-level actions into temporally extended options or
subtasks \citep{bacon2017optioncritic}. It preserves option boundaries,
aggregate effects, and sometimes duration. It may drop internal reasoning,
sub-worker identity, and cross-option dependencies, especially when the
community format assumes one active option at a time.

\subsection{Search-tree view}
\label{app:proj:mcts}

This view keeps explored states, actions, visit counts, backed-up values,
or action paths for search-style agents \citep{yao2023tree,rstarmath2025,
restmcts2024}. It supports questions about coverage, depth,
backtracking, and pruning. It may drop free-form reasoning text,
parallelism outside the search tree, and tool interactions that are not
part of the state/action abstraction.

\subsection{Full event-log view}
\label{app:proj:json-log}

The full rollout card is closest to a typed event log: it preserves
actions, observations, tool calls, artefacts, status changes, timing,
environment state, and benchmark-specific annotations. It is larger than
a reported score and less convenient than a task-specific view, but it is
the object that makes later reporting rules replayable.

\subsection{Adding a further view}
\label{app:proj:extension}

A new view can be added without changing the rollout-card principle:
publish the reporting-rule implementation or version, declare the fields and
rows it reads, and record any filters or structural collapses.
Declarative planning~\citep{ada2023llmp}, temporal planning with durative
actions and makespan~\citep{saccon2025plantor}, embodied-agent traces,
and world-model planning are natural future views. The card remains the
shared record; the reported view records one community's reporting
choice.


\section{Reference Rollout-Card Bundle Format}
\label{app:system}

\subsection{Portable archive layout}
\label{app:system:format-spec}

The rollout-card format of \S\ref{sec:system:format} is a
medium-independent bundle: any backend preserving the row
semantics below can emit and consume valid cards. A zip archive, object
storage prefix, Hugging Face dataset repository, or database export can all be
valid carriers. \textsc{Ergon} is the reference emitter and validator,
but the format does not depend on \textsc{Ergon}'s database, trainer adapters, or
runtime. The dashboard of
Figure~\ref{fig:dashboard} is a reference \emph{reader} that
loads conformant bundles. The reader is illustrative and is not part of
the format definition.

\begin{center}
\centering
\includegraphics[width=\textwidth]{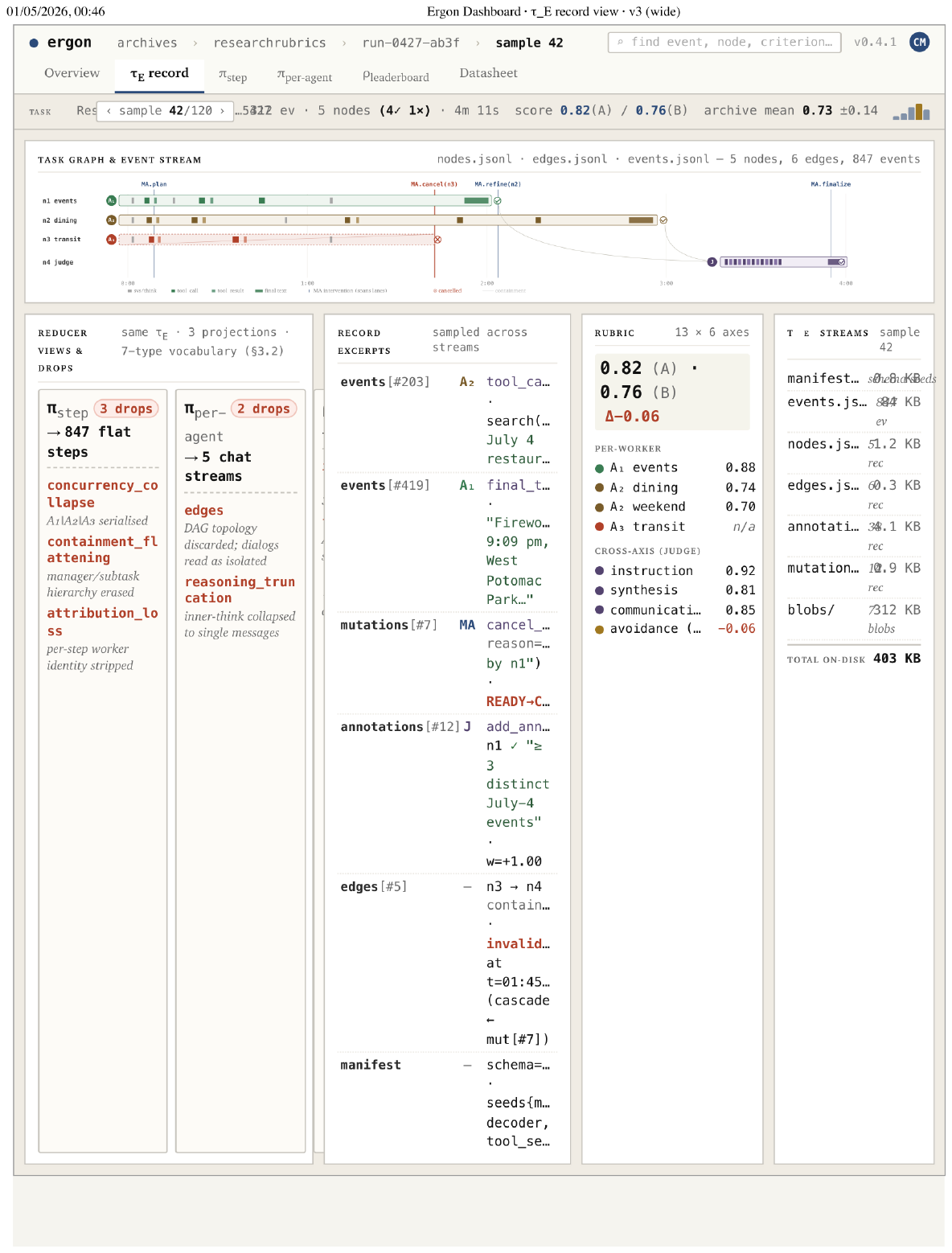}
\captionof{figure}{One sample from a rollout card, rendered by a reference
viewer over a ResearchRubrics task \citep{researchrubrics2025}. The top
panel shows the recorded run. The bottom panels show three reported views or quantities derived from
the same card: a step-indexed event view, a per-agent view, and a
leaderboard-style score. Each panel is labelled with the information
it does not carry forward.}
\label{fig:dashboard}
\end{center}

\paragraph{Bundle layout.} A card is a logical directory of
seven artefacts: \texttt{manifest.json} (run-level metadata,
format version, content hashes); five append-only JSON-lines
streams (\texttt{events.jsonl}, \texttt{nodes.jsonl},
\texttt{edges.jsonl}, \texttt{annotations.jsonl},
\texttt{mutations.jsonl}); and an optional \texttt{blobs/}
directory holding content-addressed overflow for payloads above
the inline size cap (default 64\,KB). The directory may be
distributed as-is, packed into a zip or tarball, written to an
object-storage prefix, or materialised as a Hugging Face dataset
repository with the streams as split files. Nothing in the
format assumes Postgres, Python, or any particular runtime.

\paragraph{Row schemas.} Each stream is a newline-delimited JSON file;
each row has the columns in Table~\ref{tab:rowschemas}.
Column types are JSON primitives (string, integer, object,
null); \texttt{payload} and
\texttt{old\_value}/\texttt{new\_value} columns are arbitrary
JSON objects whose shape depends on a row-level discriminator
(\texttt{event\_type} for events, \texttt{mutation\_type} for mutations).

\begin{table}[h]
\centering
\small
\caption{Rollout-card JSONL row schemas. Each stream carries a
monotonic \texttt{sequence}; its scope (per-run or
per-\texttt{task\_execution\_id}) is noted. Discriminator
columns select the \texttt{payload} shape.}
\label{tab:rowschemas}
\footnotesize
\setlength{\tabcolsep}{3pt}
\begin{tabular}{@{}l>{\raggedright\arraybackslash}p{0.72\linewidth}@{}}
\toprule
\textbf{Stream} & \textbf{Row schema} \\
\midrule
\texttt{events.jsonl}      & \texttt{(event\_id, task\_execution\_id, worker\_binding\_key, sequence, event\_type, turn\_id, payload, started\_at, completed\_at, policy\_version)}; \texttt{sequence} monotonic per \texttt{task\_execution\_id}. \\
\texttt{nodes.jsonl}       & \texttt{(node\_id, parent\_id, instance\_key, task\_key, status, assigned\_worker\_key, level, created\_at, updated\_at)}. \\
\texttt{edges.jsonl}       & \texttt{(source\_node\_id, target\_node\_id, status, created\_at, updated\_at)}; \texttt{status} $\in$ \{\texttt{pending}, \texttt{satisfied}, \texttt{invalidated}\}. \\
\texttt{annotations.jsonl} & \texttt{(target\_type, target\_id, namespace, sequence, payload, created\_at)}; latest \texttt{sequence} within \texttt{(target, namespace)} is current; prior rows retained. \\
\texttt{mutations.jsonl}   & \texttt{(sequence, mutation\_type, target\_type, target\_id, actor, old\_value, new\_value, reason, created\_at)}; \texttt{sequence} monotonic per run. \\
\bottomrule
\end{tabular}
\end{table}

\paragraph{Portability invariants.} Any backend emitting a card
must honour four invariants regardless of how rows are stored:
(i) \texttt{mutations.jsonl} is strictly append-only --- reversals appear
as new mutations, deletions as tombstones; (ii) \texttt{events.jsonl} is
append-only per \texttt{task\_execution\_id}; (iii)
\texttt{annotations.jsonl} is namespace-keyed, with the latest
\texttt{sequence} within a \texttt{(target, namespace)} pair as the
current value and prior rows retained for replay; and (iv) the dependency graph
implied by \texttt{edges.jsonl} is acyclic at every point in the run's
lifetime. These invariants are the portability contract: they make it
possible to replay a card, audit the fields used by each reporting rule, and compare views
without knowing the runtime that produced the run.

\paragraph{Extensibility.} \texttt{manifest.json} carries a
format version; consumers check the major version and tolerate
minor-version additions.
Additional columns on existing streams are permitted and must
be ignored by consumers that do not recognise them. The public
extension point for new metadata is
\texttt{annotations.jsonl}'s \texttt{namespace} field: a
consumer claims a namespace, writes domain-specific payloads
under it, and the format guarantees that round-tripping a card
through a reader that does not understand the namespace preserves those
rows intact. Namespaces can therefore carry benchmark-specific state,
domain metadata, redaction notes, licensing flags, or reporting-rule hints
without changing the base streams.

\subsection{\texorpdfstring{Reporting rules and validation}{Reporting rules and validation}}
\label{app:system:reporting-rules}
\label{app:verl}
\label{app:integrations}

A valid bundle also contains, either in \texttt{manifest.json} or in a
separate reporting-rule registry file, the reporting-rule list used for any reported
numbers. Each entry records a name, implementation or version identifier,
configuration, input streams, output target, and drops manifest. A typed
reader can generate field-level drops by logging row and field access;
semantic losses such as ``timing erased'' or ``dependency edge
collapsed'' are author declarations attached to that field-access record.

The reference implementation supplies three non-normative utilities: a
JSON-schema export for the row types in Table~\ref{tab:rowschemas}, an
exporter that writes a completed run as the archive above, and a validator
that checks layout, content hashes, monotonic sequence fields,
append-only-stream conformance, namespace consistency, and acyclicity of
the edge stream. These utilities are evidence that the format is
concrete, but any independent system can produce a conformant card by
writing the same streams and reporting-rule metadata.


\section{Experimental Setup}
\label{app:setup}

This appendix describes the source records and re-grading protocol used
in \S\ref{sec:validation}. RQ1 uses public releases that already contain
\textcolor{blue}{labels, partial logs, span/process traces, or action/search traces}; RQ2 uses public archives whose preserved outputs can be
scored under more than one reporting rule. No RQ1 result required
new policy rollouts, and no RQ2 result changes the underlying model
output, patch, submission, or trajectory being scored.

\subsection{Public source artefacts}
\label{app:actions}
\label{app:benchmarks}

Table~\ref{tab:experiment-sources} lists the public artefact classes
used in the two experiments. Appendix~\ref{app:rq1-reuse} describes the
RQ1 sources in more detail; the table here records why each source is
scoreable without rerunning the original system.

\begin{table}[h]
\centering
\small
\setlength{\tabcolsep}{4pt}
\caption{Public artefacts used for the experiments. RQ1 asks new
questions of preserved traces; RQ2 changes reporting rules while
holding the preserved artefact fixed.}
\label{tab:experiment-sources}
\begin{tabular}{@{}p{0.18\linewidth}p{0.27\linewidth}p{0.25\linewidth}p{0.22\linewidth}@{}}
\toprule
\textbf{Use} & \textbf{Source class} & \textbf{Preserved record} &
\textbf{Reporting contrast} \\
\midrule
RQ1 & GAP public results & Text-safety labels, tool-call safety labels,
forbidden-call fields & Visible response safety vs tool-channel safety \\
RQ1 & MAESTRO trace parquet & Spans, outcomes, token counts, scaffold
identifiers & Run outcome vs coordination overhead \\
RQ1 & COPRA miniF2F selected logs & Theorem outcomes, realised steps, elapsed
time, successful proof text & Proved/failed vs proof-search cost \\
RQ1 & Tree-of-Thought crossword logs & Paired prune/no-prune DFS action
paths and reward snapshots & Final reward vs search shape \\
RQ2 & Public answer, patch, submission, and trajectory archives &
Archived artefacts for the same model output or benchmark case &
Alternative graders, denominators, thresholds, weights, and trajectory
success definitions \\
\bottomrule
\end{tabular}
\end{table}

\textcolor{blue}{The public Hugging Face release contains 21 card exports: 17 trace-publication exports plus four analytic/recovered-view non-trace exports. The 17 trace-publication exports consist of 11 full multi-step traces, one span/process trace export, and five one-step trace exports. The four non-trace exports are GAP, COPRA miniF2F, MLE-Bench, and SWE-bench cross-harness. These four exports preserve labels, partial logs, submissions, verdicts, view outputs, reporting-rule metadata, and drops manifests sufficient for the paper's reanalyses, but they are not full rollout traces and we do not claim that they support open-ended community reanalysis in the same way as the trace-publication exports. The broader 25-source corpus remains available only through the local artefact build rather than the public release. Table~\ref{tab:artifact-corpus} groups the released exports by evidence type.}

\begin{table}[h]
\centering
\small
\setlength{\tabcolsep}{4pt}
\caption{Preserved evidence classes in the \textcolor{blue}{released rollout-card exports}. The
machine-readable \textcolor{blue}{artefact} records the per-source field inventory, views,
reporting rules, and drops manifests.}
\label{tab:artifact-corpus}
\begin{tabular}{@{}p{0.23\linewidth}p{0.36\linewidth}p{0.33\linewidth}@{}}
\toprule
\textbf{Evidence class} & \textbf{Examples} & \textbf{\textcolor{blue}{Artefact treatment}} \\
\midrule
\textcolor{blue}{Full multi-step traces} & \textcolor{blue}{AgentRewardBench, AgentHarm, ATBench, Debate/MALLM, MiniWoB++, StableToolBench, SWE-smith, \(\tau\)-bench, Tree-of-Thought Crosswords, Tree-of-Thought Game24, WebLINX} & \textcolor{blue}{Trace-publication exports with inspectable messages, tool calls, browser/UI actions, or search steps} \\
\textcolor{blue}{Span/process traces} & \textcolor{blue}{MAESTRO} & \textcolor{blue}{Trace-publication export with workers, durations, statuses, token counts, and run-level process evidence; not a full action-level transcript} \\
\textcolor{blue}{One-step traces} & \textcolor{blue}{BFCL, BrowseComp, GPQA, HumanEval, MMLU} & \textcolor{blue}{Trace-publication exports with prompt, response or function-call, and outcome records; not full rollout traces} \\
\textcolor{blue}{Analytic/recovered-view non-trace exports} & \textcolor{blue}{GAP, COPRA miniF2F, MLE-Bench, SWE-bench cross-harness} & \textcolor{blue}{Labels, partial logs, submissions, verdicts, view outputs, reporting-rule metadata, and drops manifests used for paper reanalysis; not full rollout traces and not claimed as reusable trace datasets} \\
\bottomrule
\end{tabular}
\end{table}

\subsection{\texorpdfstring{\textcolor{blue}{Storage profile of released card exports}}{Storage profile of released card exports}}
\label{app:export-storage-profile}

\textcolor{blue}{The public export metadata is sufficient to profile storage overhead, although it does not retain wall-clock exporter or validator timings. Table~\ref{tab:export-storage-profile} reports representative released exports using Hugging Face file manifests. The total size includes sharded run records, reducer outputs, drops manifests, resource blobs, checksums, and manifest files; the compression ratio is the uncompressed-to-compressed size ratio for the parquet shards only. Resource-heavy span traces can be much larger than one-step or analytic exports, but the compact parquet streams compress by roughly \(3\)--\(5\times\) for the larger releases.}

\begin{table}[h]
\centering
\scriptsize
\setlength{\tabcolsep}{3pt}
\caption[Storage profile of representative released card exports]{\textcolor{blue}{Storage profile of representative released card exports. Sizes are computed from the Hugging Face file manifest for each export; parquet compression excludes JSON resource blobs.}}
\label{tab:export-storage-profile}
\begin{tabular}{@{}p{0.18\linewidth}p{0.22\linewidth}r r r r@{}}
\toprule
\textbf{\textcolor{blue}{Export}} & \textbf{\textcolor{blue}{Evidence class}} & \textbf{\textcolor{blue}{Runs}} & \textbf{\textcolor{blue}{Total MB}} & \textbf{\textcolor{blue}{KB/run}} & \textbf{\textcolor{blue}{Parquet comp.}} \\
\midrule
\textcolor{blue}{MAESTRO} & \textcolor{blue}{Span/process trace} & \textcolor{blue}{1{,}056} & \textcolor{blue}{4{,}250.4} & \textcolor{blue}{4{,}025.0} & \textcolor{blue}{3.4$\times$} \\
\textcolor{blue}{\(\tau\)-bench} & \textcolor{blue}{Multi-step tool trace} & \textcolor{blue}{1{,}980} & \textcolor{blue}{179.8} & \textcolor{blue}{90.8} & \textcolor{blue}{4.5$\times$} \\
\textcolor{blue}{MMLU} & \textcolor{blue}{One-step trace} & \textcolor{blue}{28{,}084} & \textcolor{blue}{51.4} & \textcolor{blue}{1.8} & \textcolor{blue}{4.1$\times$} \\
\textcolor{blue}{GPQA} & \textcolor{blue}{One-step trace} & \textcolor{blue}{600} & \textcolor{blue}{3.2} & \textcolor{blue}{5.3} & \textcolor{blue}{3.6$\times$} \\
\textcolor{blue}{GAP} & \textcolor{blue}{Recovered non-trace view} & \textcolor{blue}{17{,}420} & \textcolor{blue}{30.7} & \textcolor{blue}{1.8} & \textcolor{blue}{4.2$\times$} \\
\textcolor{blue}{MLE-Bench} & \textcolor{blue}{Recovered non-trace view} & \textcolor{blue}{14{,}466} & \textcolor{blue}{30.4} & \textcolor{blue}{2.1} & \textcolor{blue}{3.5$\times$} \\
\textcolor{blue}{Tree-of-Thought Crosswords} & \textcolor{blue}{Search trace} & \textcolor{blue}{40} & \textcolor{blue}{0.16} & \textcolor{blue}{4.1} & \textcolor{blue}{3.0$\times$} \\
\bottomrule
\end{tabular}
\end{table}

\subsection{\texorpdfstring{RQ1 dataset conversion and recovered views}{RQ1 dataset conversion and recovered views}}
\label{app:role-diversity}
\label{app:fivescaffolds}

Each RQ1 source is converted into the rollout-card representation at the
level of fields available in the public release. We assign stable run
identifiers, preserve source-level row or span identifiers, and record
source-specific fields under a namespace for that benchmark.
The recovered view is then a deterministic analysis over those preserved
fields. For example, GAP rows preserve both \texttt{t\_safe} and
\texttt{tc\_safe}; MAESTRO rows preserve span metadata and token counts;
COPRA miniF2F records preserve theorem-level outcome and realised search cost;
and Tree-of-Thought crossword logs preserve action-list prefixes and
reward snapshots. Appendix~\ref{app:rq1-reuse} gives the per-case
source files, extraction rules, and missing-field limitations.

The recovered cards are intentionally conservative. We do not infer
unobserved failed proof branches for the COPRA miniF2F logs, recover hidden pruning
decisions for Tree-of-Thought, assign causal labels for MAESTRO
coordination overhead, or reconstruct the full GAP runtime environment.
Those absent fields are limitations of the recovered view and are treated
as drops from the source artefact.

These experiments evaluate a publication practice, not a new task
leaderboard. RQ1 findings are descriptive reanalyses of preserved public
records: they do not estimate deployment prevalence or prove that process
variables such as coordination overhead caused failure. RQ2 holds the
preserved artefact fixed and changes reporting rules, so it measures
convention sensitivity rather than the intrinsic quality of a benchmark
or model.


\subsection{Cross-harness reconciliation methodology}
\label{app:reconciliation}

The re-grading sweep in \S\ref{sec:validation:results} uses only
archives that satisfy a counterfactual-scoreability criterion: the
public artefact must preserve enough of the model's output to evaluate
the same run under at least two plausible reporting rules. This
excludes papers that publish only aggregate scores. It includes final
answers for short-answer tasks, submitted patches for software
engineering tasks, Kaggle submissions for MLE-Bench-style tasks, and
tool trajectories or database states for long-horizon interaction
benchmarks.

For each benchmark family, we write a dataset adapter that converts
the public artefact into a rollout-card record with a corresponding
drops manifest. The manifest records information that the public
archive did not preserve, information that is implicit in the original
harness, and fields that can be recovered only by adopting a
benchmark-specific convention. We then run multiple reporting
functions over the fixed record. Depending on the benchmark, those
functions vary answer extraction, LLM-judge or rule-based grading,
failure handling, threshold or medal assignment, slice weighting, or
trajectory success definitions.

We report the \textcolor{blue}{score gap} that the reporting function induces, not a new
preferred leaderboard score. For each archive we
record \textcolor{blue}{score gaps} under alternative conventions, ranking inversions
where model order changes, and calibration cases where independent
reporting functions agree. In the SWE-bench Verified case, the
principal failure-handling contrast is whether no-submission outcomes
are included as unresolved attempts or excluded from the scored subset:
SWE-agent has 50 such outcomes among 500 public runs, while Agentless
has 4 among 500. These two conventions bracket a common harness-level
choice and make the convention-dependent share of the \textcolor{blue}{reported score gap}
explicit. LLM-judge variance is one source of reporting \textcolor{blue}{score gaps}, but the
re-grading sweep also includes deterministic conventions such as
denominator choice, thresholding, slice weighting, and extraction.

\section*{NeurIPS Paper Checklist}

\begin{enumerate}

\item {\bf Claims}
    \item[] Question: Do the main claims made in the abstract and introduction accurately reflect the paper's contributions and scope?
    \item[] Answer: \answerYes{}
    \item[] Justification: The abstract and \S\ref{sec:intro} claim (i) that current agent research publishes reported numbers without rollouts, producing cross-community fragmentation and reporting-convention variance, and (ii) that rollout cards plus drops manifests address both problems; these are substantiated by the 50-repo survey (Appendix~\ref{app:survey}), the 37-pair variance catalogue (Appendix~\ref{app:variance-catalogue}), the format specification in \S\ref{sec:system} and Appendix~\ref{app:system}, and the two retrospective experiments of \S\ref{sec:validation}. The accompanying \textcolor{blue}{artefact} and public Hugging Face exports make the release concrete: 21 card exports are published, comprising 17 trace-publication exports plus four analytic/recovered-view non-trace exports that support the paper reanalyses. Scope limits are explicit: the evidence covers 50 audited repositories, 37 documented variance pairs, four RQ1 public-rollout reuse cases, public artefacts that preserve enough information for re-grading, and the 21 public card exports just described; it is not a prevalence estimate over all agent releases.

\item {\bf Limitations}
    \item[] Question: Does the paper discuss the limitations of the work performed by the authors?
    \item[] Answer: \answerYes{}
    \item[] Justification: \S\ref{sec:conclusion} states that rollout cards do not choose a canonical metric and do not remove the need for judgment about privacy, licensing, redaction, benchmark design, metric choice, or selective reporting. It also scopes the evidence to 50 audited repositories, 37 documented variance pairs, four public-rollout reuse cases, and public artefacts rich enough for reanalysis or re-grading. Appendix~\ref{app:rq1-reuse} records missing fields for the recovered RQ1 views, and Appendix~\ref{app:setup} distinguishes the 17 trace-publication exports from the four analytic/recovered-view non-trace exports.

\item {\bf Theory assumptions and proofs}
    \item[] Question: For each theoretical result, does the paper provide the full set of assumptions and a complete (and correct) proof?
    \item[] Answer: \answerNA{}
    \item[] Justification: The paper does not state theorems or prove formal results. Appendix~\ref{app:projections} gives illustrative preservation examples for views and reporting rules; these are explanatory examples of what fields a view preserves or drops, not theorem statements or proof obligations.

    \item {\bf Experimental result reproducibility}
    \item[] Question: Does the paper fully disclose all the information needed to reproduce the main experimental results of the paper to the extent that it affects the main claims and/or conclusions of the paper (regardless of whether the code and data are provided or not)?
    \item[] Answer: \answerYes{}
    \item[] Justification: \S\ref{sec:validation:setup} describes the public source records and re-grading protocol. Appendix~\ref{app:rq1-reuse} gives the RQ1 source artefacts, extraction rules, and missing-field limits; Appendix~\ref{app:setup} gives the public artefact classes, ingestion rules, and reporting-convention reconciliation method; Appendix~\ref{app:system} specifies the portable rollout-card bundle format. The RQ1/RQ2 reported numbers are deterministic functions of preserved public records, source artefacts, or declared re-grading conventions. The \textcolor{blue}{artefact} provides a default Hugging Face-backed reconstruction path for public card exports, plus a heavier local Ergon rebuild path for provenance depth; the 50-repo survey counts are released separately as verifiable audit records with pinned source references.

\item {\bf Open access to data and code}
    \item[] Question: Does the paper provide open access to the data and code, with sufficient instructions to faithfully reproduce the main experimental results, as described in supplemental material?
    \item[] Answer: \answerYes{}
    \item[] Justification: An anonymised artefact bundle accompanies this submission, including the rollout-card format specification, reference implementation utilities, dataset adapters, derived rollout-card exports, drops manifests, and analysis scripts. The public Hugging Face release is hosted under the anonymous namespace \url{https://huggingface.co/annon124816} and consists of 21 card exports: 17 trace-publication exports plus four analytic/recovered-view non-trace exports. The 17 trace-publication exports are \texttt{agent\_reward\_bench}, \texttt{agentharm}, \texttt{atbench}, \texttt{bfcl}, \texttt{browsecomp}, \texttt{debate\_mallm}, \texttt{gpqa}, \texttt{humaneval}, \texttt{maestro}, \texttt{miniwob}, \texttt{mmlu}, \texttt{stabletoolbench}, \texttt{swe\_smith}, \texttt{tau\_bench}, \texttt{tot\_crosswords}, \texttt{tot\_game24}, and \texttt{weblinx}. The four non-trace exports are \texttt{gap}, \texttt{copra}, \texttt{mle\_bench}, and \texttt{swebench\_cross\_harness}. Each export includes a dataset README and metadata files, including \texttt{dataset\_info.json}; the OpenReview Croissant package contains RAI-augmented Croissant metadata for all 21 exports. The public releases used in RQ1 and RQ2 are cited in \S\ref{sec:validation}; extraction and reconciliation protocols are specified in Appendix~\ref{app:rq1-reuse} and Appendix~\ref{app:reconciliation}.

\item {\bf Experimental setting/details}
    \item[] Question: Does the paper specify all the training and test details (e.g., data splits, hyperparameters, how they were chosen, type of optimizer) necessary to understand the results?
    \item[] Answer: \answerYes{}
    \item[] Justification: The paper reports no new model training or new policy rollouts. RQ1 reanalyses preserved public process records, and RQ2 re-grades preserved public answers, patches, submissions, or trajectories under declared alternative conventions. Appendix~\ref{app:setup} specifies the source artefact classes, ingestion rules, and convention choices used for the re-grading sweep. The released card datasets are derived publication artefacts over public sources rather than new agent executions.

\item {\bf Experiment statistical significance}
    \item[] Question: Does the paper report error bars suitably and correctly defined or other appropriate information about the statistical significance of the experiments?
    \item[] Answer: \answerYes{}
    \item[] Justification: The RQ2 reporting-convention reconciliation is a deterministic re-grading of already-released artefacts, so no stochastic variation enters the pipeline; we report exact denominators, score deltas, and ranking inversions rather than error bars. For RQ1, the figure reports descriptive summaries over preserved public records and sample counts for binned summaries; the COPRA panel additionally reports an observational logistic trend with an explicit causal caveat.

\item {\bf Experiments compute resources}
    \item[] Question: For each experiment, does the paper provide sufficient information on the computer resources (type of compute workers, memory, time of execution) needed to reproduce the experiments?
    \item[] Answer: \answerYes{}
    \item[] Justification: No model training and no new policy rollouts are performed. RQ1 consists of local analysis over public trace artefacts; RQ2 consists of ingestion and deterministic re-grading of preserved public artefacts. The default Hugging Face-backed reconstruction path downloads already-exported cards and runs local analytics; the heavier raw-source rebuild ingests public artefacts through Ergon/Postgres and may require additional storage for large trace resources. No GPU is required for any reported result.

\item {\bf Code of ethics}
    \item[] Question: Does the research conducted in the paper conform, in every respect, with the NeurIPS Code of Ethics \url{https://neurips.cc/public/EthicsGuidelines}?
    \item[] Answer: \answerYes{}
    \item[] Justification: The paper proposes a publication format, reference implementation, and public card exports for agent research. It involves no human subjects and no private data collection. The analyses and released card exports are derived from public benchmark, repository, dataset, trajectory, patch, submission, or fixed-output artefacts, with provenance, licensing, and redistribution notes recorded in the \textcolor{blue}{artefact}. The reference implementation is released as open source.

\item {\bf Broader impacts}
    \item[] Question: Does the paper discuss both potential positive societal impacts and negative societal impacts of the work performed?
    \item[] Answer: \answerYes{}
    \item[] Justification: Positive impacts --- improved auditability of agent-research claims, reporting-convention reconciliation, and legibility of methodology differences --- are discussed in \S\ref{sec:conclusion}. The main risk is that rollout records and public card exports can contain prompts, model outputs, tool traces, or environment data requiring privacy, licensing, redaction, or access-control decisions; \S\ref{sec:conclusion} states that rollout cards make such choices inspectable rather than eliminating them.

\item {\bf Safeguards}
    \item[] Question: Does the paper describe safeguards that have been put in place for responsible release of data or models that have a high risk for misuse (e.g., pre-trained language models, image generators, or scraped datasets)?
    \item[] Answer: \answerYes{}
    \item[] Justification: The paper releases a format specification (rollout cards), reference implementation utilities, and derived card datasets, not a pre-trained model or generative system. Section~\ref{sec:system} treats release-scope metadata and drops manifests as part of the rollout-card standard: card exports declare redaction, access, licensing, redistribution, omission, and source-policy limits rather than requiring every source field to be publicly redistributed. The 21 public Hugging Face exports are derived from public upstream artefacts; the artefact records license notes and redistribution limits; raw upstream datasets are omitted where redistribution is not permitted; packaging applies \textcolor{blue}{basic credential/secret redaction}; and the release distinguishes trace-publication exports from analytic/recovered-view non-trace exports when full traces are not published.

\item {\bf Licenses for existing assets}
    \item[] Question: Are the creators or original owners of assets (e.g., code, data, models), used in the paper, properly credited and are the license and terms of use explicitly mentioned and properly respected?
    \item[] Answer: \answerYes{}
    \item[] Justification: The paper cites the public sources used for RQ1/RQ2 analysis and card export in \S\ref{sec:validation}, \S\ref{sec:related}, Appendix~\ref{app:rq1-reuse}, and Appendix~\ref{app:reconciliation}. The \textcolor{blue}{artefact} records per-source URLs, version references or checksums where available, license notes, and redistribution limits for the 21 public Hugging Face exports and the broader local card corpus. Source datasets themselves are not re-released as source datasets unless the source policy permits derived public redistribution; the released objects are derived rollout-card exports with provenance and drops metadata.

\item {\bf New assets}
    \item[] Question: Are new assets introduced in the paper well documented and is the documentation provided alongside the assets?
    \item[] Answer: \answerYes{}
    \item[] Justification: New assets accompany the paper: (i) the rollout-card format specification (Appendix~\ref{app:system}); (ii) reference exporter, validator, and reader utilities for conformant bundles; (iii) dataset adapters, view/reporting-rule tooling, and drops-manifest tooling; (iv) 21 public Hugging Face card exports, comprising 17 trace-publication exports plus four analytic/recovered-view non-trace exports; and (v) ingestion, re-grading, and reporting-rule scripts for the public artefacts used in RQ1 and RQ2. The public dataset exports include dataset READMEs and machine-readable metadata; the OpenReview Croissant package contains RAI-augmented Croissant metadata for all 21 exports. The repository also documents a broader 25-source local card corpus, but the public-upload claim is the 21-export release described above. All new assets are documented in-repository and anonymised for submission.

\item {\bf Crowdsourcing and research with human subjects}
    \item[] Question: For crowdsourcing experiments and research with human subjects, does the paper include the full text of instructions given to participants and screenshots, if applicable, as well as details about compensation (if any)?
    \item[] Answer: \answerNA{}
    \item[] Justification: The paper involves no crowdsourcing and no human subjects. The public-rollout reanalyses, 50-repo survey, and 37-pair variance catalogue are authored by the paper's authors against public artefacts.

\item {\bf Institutional review board (IRB) approvals or equivalent for research with human subjects}
    \item[] Question: Does the paper describe potential risks incurred by study participants, whether such risks were disclosed to the subjects, and whether Institutional Review Board (IRB) approvals (or an equivalent approval/review based on the requirements of your country or institution) were obtained?
    \item[] Answer: \answerNA{}
    \item[] Justification: No human subjects research is performed. No IRB review is applicable.

\item {\bf Declaration of LLM usage}
    \item[] Question: Does the paper describe the usage of LLMs if it is an important, original, or non-standard component of the core methods in this research? Note that if the LLM is used only for writing, editing, or formatting purposes and does \emph{not} impact the core methodology, scientific rigor, or originality of the research, declaration is not required.
    \item[] Answer: \answerYes{}
    \item[] Justification: LLMs and LLM agents are the central object of study. RQ1 reanalyses preserved public traces from tool-using, multi-agent, theorem-proving, and tree-search systems; RQ2 re-grades preserved outputs and trajectories produced by LLM systems under alternative reporting conventions. No LLM is used as a grading, inference, or decision component for the repository audit. That survey is documented as a pinned public-source audit: Appendix~\ref{app:survey} describes the rubric and acceptance criteria, and the supplementary survey bundle releases final labels, evidence snippets, and pinned source references so the survey counts are verifiable from public audit records.

\end{enumerate}

\end{document}